%% file: main.tex
\documentclass[lettersize,journal]{IEEEtran}
\usepackage{amsmath,amsfonts,amssymb,array}
\usepackage[dvipsnames]{xcolor}
\usepackage{textcomp}
\usepackage{stfloats}
\usepackage{url}
\usepackage{verbatim}
\usepackage{graphicx}
\usepackage{subfigure}
\usepackage[numbers,sort&compress]{natbib}
\usepackage{booktabs}
\usepackage{overpic}
\usepackage{multirow}
\usepackage[ruled,linesnumbered]{algorithm2e}
\usepackage[colorlinks]{hyperref}
\usepackage[figuresright]{rotating}
\newcommand{\eg}{e.g.}
  
\newcommand{\fb}[2][\textbf{FRANCESCO: }]{{\leavevmode\color{black}#1#2}}

\newcommand{\um}[1]{{\leavevmode\color{black} #1}}

\newcommand{\ie}{i.e.}

\newcommand{\notation}[1]{#1}

\DeclareMathOperator*{\argmax}{arg\,max}

\begin{document}

\title{RECALL+: Adversarial Web-based Replay for Continual Learning in Semantic Segmentation}

\author{Chang Liu{$^{1,2}$,}
Giulia Rizzoli{$^{2}$},~\IEEEmembership{Student Member,~IEEE,}
Francesco Barbato{$^{2}$},~\IEEEmembership{Student Member,~IEEE, } \\
Andrea Maracani{$^{2}$}, 
Marco Toldo{$^{2}$},
Umberto Michieli{$^{2}$},
Yi Niu{$^{1,3}$},
Pietro Zanuttigh{$^{2}$},~\IEEEmembership{Member,~IEEE}

$^{1}$\,School of Artificial Intelligence,
Xidian University, China \\
$^{2}$\,Department of Information Engineering,
University of Padova, Italy \\
$^{3}$\,The Pengcheng Lab, China\\

\thanks{
This work is partially funded by the China Scholarship Council and by the SEED project Semantic Segmentation in the Wild.}}

\markboth{Journal of \LaTeX\ Class Files,~Vol.~14, No.~8, August~2021}%
{Shell \MakeLowercase{\textit{et al.}}: A Sample Article Using IEEEtran.cls for IEEE Journals}


\maketitle

\begin{abstract}
Catastrophic forgetting of previous knowledge is a critical issue in continual learning typically handled through various regularization strategies. However, existing methods struggle especially when several incremental steps are performed. In this paper, we extend our previous approach (RECALL) and tackle forgetting by exploiting unsupervised web-crawled data to retrieve examples of old classes from online databases.
In contrast to the original methodology, which did not incorporate an assessment of web-based data, the present work proposes two advanced techniques: an adversarial approach and an adaptive threshold strategy. These methods are utilized to meticulously choose samples from web data that exhibit strong statistical congruence with the no longer available training data.
Furthermore, we improved the pseudo-labeling scheme to achieve a more accurate labeling of web data that also considers classes being learned in the current step. Experimental results show that \um{this enhanced} approach achieves \um{remarkable} results, particularly when the incremental scenario spans multiple steps.
\end{abstract}


\section{Introduction}
\label{sec:intro}

Continual learning strategies allow machine learning models to be trained incrementally over multiple steps instead of relying on a single-step training on a large dataset \cite{parisi2019continual}. This capability is crucial in practical applications where privacy  or intellectual property concerns may render the original training data unavailable when new tasks are introduced.
The problem has been widely investigated for image classification and several approaches have been proposed to address the challenges associated with learning new classes without compromising the recognition of already learned ones \cite{rebuffi2017icarl,douillard2022dytox,kirkpatrick2017overcoming}. Particularly, when a model is forced to learn a new task without additional constraints, the optimization will result in the so-called catastrophic forgetting phenomenon: the model tends to overfit the new data, forgetting the knowledge of old concepts.
In recent years, the problem has been investigated in more challenging tasks, such as semantic segmentation. Standard approaches for class-incremental semantic segmentation take inspiration from classification works, extending knowledge-distillation objectives to dense predictions \cite{yang2022uncertainty,shang2023incrementer}. While such approaches can alleviate the rate at which old knowledge is lost, they face issues when multiple incremental steps are considered or when the semantic content of the classes varies.

In our conference paper (RECALL,  Replay-based continual learning in semantic segmentation \cite{maracani2021recall}), we approached the problem from a unique perspective. Rather than relying on additional losses or regularization strategies, our proposal involved generating examples of old classes through generative methods or utilizing web-crawled data. This framework effectively resulted in an exemplar-free replay-based solution.
Considering that web examples lack ground-truth labels, we generated pseudo-labels using a side labeling module, which requires minimal extra storage. Besides, a self-inpainting strategy was adopted to reduce background shift by re-assigning the background region with predictions of the previous model.

\begin{figure}[t]
\includegraphics[width=\linewidth]{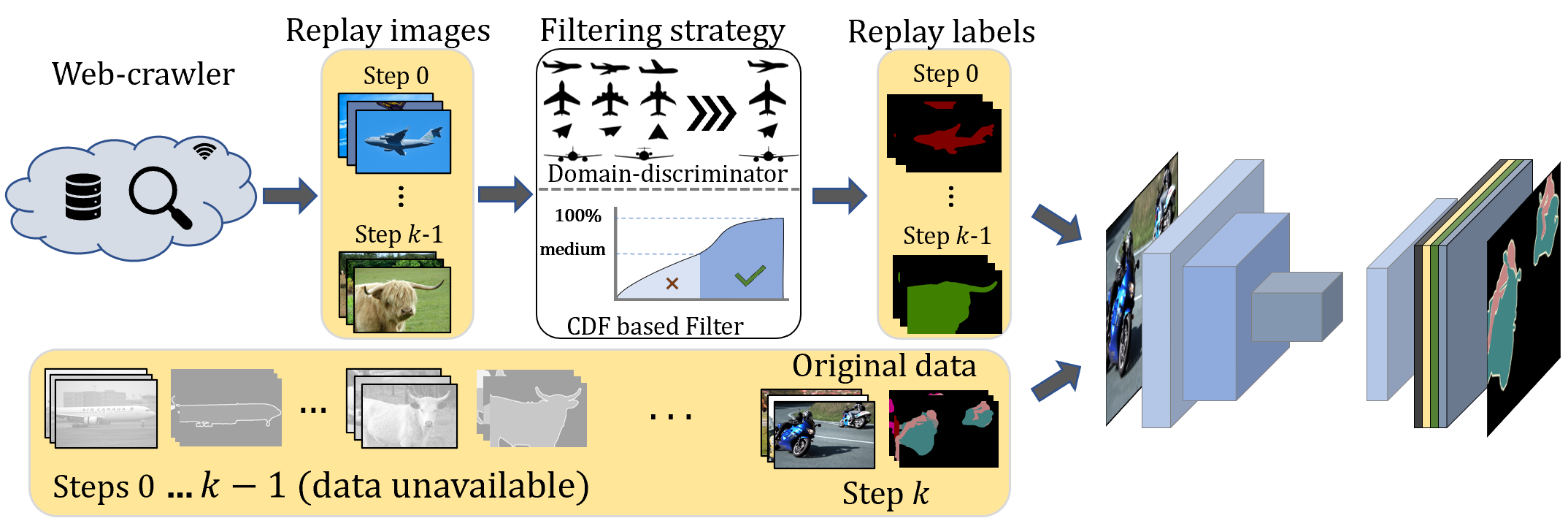}
\caption{Replay images of previously seen classes are retrieved by a web crawler and then filtered by a domain discriminator, after which the network is incrementally trained with a mixture of new and replay data.
}\vspace{-0.35cm}
\label{fig:GA}
\end{figure}

\begin{figure}[t]
\centering
\vspace{-0.2cm}
\includegraphics[trim=3cm 11.7cm 23cm 3cm, clip, width=\linewidth]{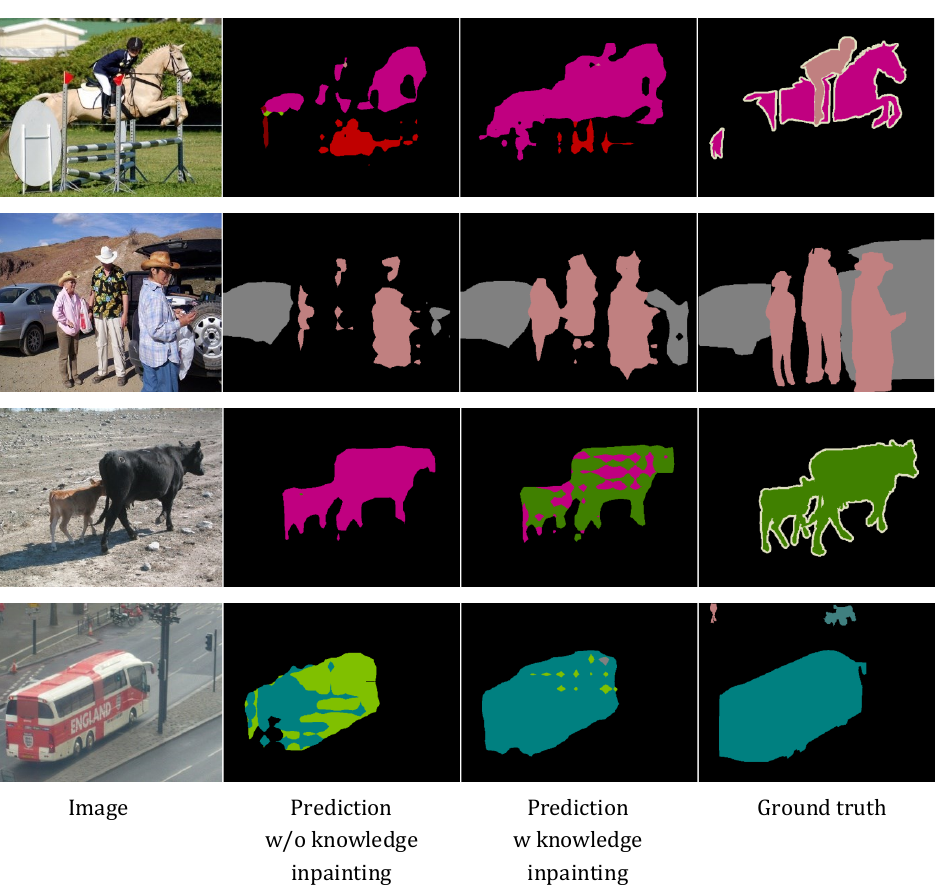}
\vspace{-0.2cm}
\caption{Comparison of predictions between the model trained with knowledge self-inpainting and without inpainting. Inpainting the replay images accelerates convergence and improves segmentation accuracy. 
} 
\label{fig:PvsWP}
\vspace{-0.35cm}
\end{figure}

 A significant limitation of our original methodology lies in the replay-data selection process, which did not ensure that web replay samples were closely aligned with the previous data distribution.
To further improve \um{performance} and strengthen the control over the web-crawled data, in this paper, we propose an image selection technique that combines an adversarial and a thresholding strategy to obtain images as close as possible to the original training dataset distribution.
In particular, we trained a discriminator network to distinguish between in-distribution and out-of-distribution samples and use it to preserve only the images that are able to fool it. The thresholding strategy, instead, is based on the pixel-class distribution, computed from the \um{training dataset} ground truth maps.
On top of the filtering strategies\um{,} we also introduce a refined inpainting strategy, where the knowledge of new 
classes is propagated to the replay samples, reducing significantly the \textit{background} shift. A few examples showing its effectiveness are  reported in Figure \ref{fig:PvsWP}.


Our contributions can be summarized as:
1) we present RECALL+, an exemplar-free replay-based approach for continual semantic segmentation that leverages web-crawled images; 
2) we propose a selection technique for web images that 
reduces the domain gap with real-world images and no-longer-available training data;
3) we devise a knowledge inpainting strategy that allows to introduce current task information in the the replay images;
4) the proposed approach achieves state-of-the-art results in a wide range of scenarios, especially when performing multiple incremental steps.

\um{We remark} that the conference version of this work did not include the selection strategy for the web-crawled images and the information \um{related to} current tasks 
in the replay images was not exploited. This extended version \um{addresses these limitations by: designing a combined (i) adversarial learning and (ii) thresholding scheme for image selection, (iii) by proposing a new knowledge-updating strategy and finally (iv) evaluates its effectiveness in a wider range of settings}. 

\section{Related Works}
\label{sec:related}


\textbf{Continual Learning (CL).} Many different techniques have been proposed to tackle catastrophic forgetting in continual learning. 
The first possible strategy is to use dynamic architectures, both allowing 
the growth of new network branches during the incremental steps \cite{wang2017growing,li2018learning}, or assuming that some network weights are available for certain tasks only \cite{fernando2017pathnet,serra2018overcoming, zhang2022representation}.
Some approaches introduce additional loss terms to regularize training  \cite{kirkpatrick2017overcoming,zenke2017continual} or distill knowledge from the model at previous steps \cite{shmelkov2017incremental,li2018learning,michieli2021knowledge, michieli2021continual}. 
These regularization-based strategies are often complementary to rehearsal-based strategies and can be used in conjunction to enhance the learning process.

The task becomes simpler if we relax the assumption that no previous samples can be used: rehearsal-based approaches store a set of samples of past tasks that can be exploited to stabilize the training \cite{rebuffi2017icarl,lopez2017gradient}. 
While storing past samples in a replay buffer may require significant memory resources. Besides, if the replay buffer contains a biased or limited set of samples, the agent may overfit these samples and fail to generalize well to unseen situations.
A \um{viable} solution to exploit rehearsal without storing previous samples is to rely on generative models, such as GANs  \cite{shin2017continual,wu2018incremental,he2018exemplar} or variational auto-encoders (VAEs)~\cite{kamra2017deep} 
to generate artificial samples. 

This work can be considered a further insight into the replay strategy: we adopt web-crawled images to prevent forgetting, avoiding both storing samples to preserve previous knowledge \cite{rebuffi2017icarl,lopez2017gradient}, and the usage of trained generative models to synthesize images \cite{shin2017continual,he2018exemplar,wu2018incremental}, thus reducing memory and computation time.

\textbf{CL in Semantic Segmentation.}
Class-incremental continual learning has been widely studied in the image classification field, while only recently it has been tackled also in the more challenging semantic segmentation task \cite{michieli2019incremental}.
Early approaches utilized knowledge distillation and regularization techniques such as parameters freezing or class re-weighting schemes \cite{michieli2019incremental,michieli2021knowledge,klingner2020class,cermelli2020modeling} to ensure the introduction of new classes while preserving the knowledge from previously learned ones.
Subsequently, regularization and contrastive mechanisms at the feature level were explored to improve class-conditional capabilities and the preservation of spatial relationships \cite{michieli2021continual,douillard2021plop,yang2022uncertainty,toldo2022learning}.
Further research was employed in the proposal of new objective functions suited for CL, among which: a class-similarity-weighted loss function to relate new classes with previously seen ones \cite{phan2022class}; a structure-preserving loss to maintain the discriminative ability of previous classes \cite{lin2022continual}; a biased-context-insensitive consistency loss that rectifies the context of old classes with respect to new ones \cite{zhao2022rbc}.
Moreover, similar to classification tasks, dynamic architecture methods have been proposed for CL in semantic segmentation: \cite{zhang2022representation} incorporates model compensation with a re-parameterization technique to preserve model complexity; \cite{xiao2023endpoints} embeds a dynamic balance parameter calculated by the ratio of new classes to merge frozen and trained branches.
In addition to the aforementioned advancements, CL in semantic segmentation has also extended to other expanding fields, such as weakly supervised segmentation \cite{cermelli2022incremental, yu2023foundation}, leveraging CL through transformers \cite{shang2023incrementer, cermelli2023comformer}, and exploring CL within distributed learning frameworks \cite{dong2023federated}.

\textbf{Webly-Supervised Learning} 
is a new research direction where large amounts of web data are used to train deep learning models \cite{chen2015webly,divvala2014learning,niu2018webly}.
It has also been employed for semantic segmentation, however, a critical challenge is that image \cite{jin2017webly,shen2018bootstrapping, yu2022self} and video \cite{hong2017weakly} data from the web comes with only weak image-level class labels, while the pixel-level semantic labelling is missing.
Current research directions include understanding how to query and select images \cite{duan2020omni} and how to exploit weakly supervised data \cite{luo2020webly,yang2020webly, yu2022self} (\eg, computing pseudo-labels). 
Recently, Yu et. al\cite{yu2022self} introduced a continual learning approach for image semantic segmentation that stores the model's knowledge before each new task and later refines pseudo-labels through a Conflict Reduction Module after training on new data.
To our knowledge, however, the first approach exploiting web data in continual learning as a replay strategy is the conference version of this work \cite{maracani2021recall}.

\section{Problem Formulation}
\label{sec:problem}
This work focuses on the semantic segmentation task, \ie, pixel-wise labeling of an  image.
Formally, given a class set $\mathcal{C} = \{b, c_1, \dots, c_{C-1}\}$ containing $C$ classes, where we assume that class $0$ is  the ``special'' \textit{background} class $b$, the task is to process an image $\mathbf{X} \in \mathcal{X} \subset \mathbb{R}^{H \times W \times 3}$ with a deep learning model \um{$M$} (typically an encoder-decoder architecture, i.e, $M  = \! D  \circ  E$) to produce a segmentation map $\mathbf{Y^{max}} \in \mathcal{Y} \subset \mathcal{C}^{H \times W}$. 

%

In the standard \um{supervised deep learning} setting, the model is tuned in a single stage on a training set $\mathcal{T} \subset \mathcal{X} \times \mathcal{Y}$, where all classes and samples are available.
In the class-incremental setting, instead, the training is assumed to happen over a sequence of \textit{steps} $k = 0, \dots, K$.
At each step, we have access to a training set $\mathcal{T}_k$ where the pixels are annotated only with the classes in a subset $\mathcal{C}_k \subset \mathcal{C}$, while pixels not belonging to $\mathcal{C}_k$ categories are labeled with \textit{background} class that, by default, it is included only in $\mathcal{C}_{0}$.
Notice that in the incremental setting the set of classes are disjoint, i.e., $\mathcal{C}_{0\rightarrow (k-1)} \cap \mathcal{C}_k = \varnothing$, where $\mathcal{C}_{0\rightarrow (k-1)} := \bigcup_{i=0}^{k-1} \mathcal{C}_i$.
We considered two standard  settings \cite{michieli2019incremental,michieli2021continual,cermelli2020modeling}: \textit{disjoint} and \textit{overlapped}.
In both settings, only the pixels belonging to the classes of the current task $\mathcal{T}_k \subset \mathcal{T}$ are labeled (previous and unseen classes are masked as \textit{background}). Nonetheless, in the former scenario, images of the current task contain only pixels from the current and the old sets of classes $\mathcal{C}_{0 \rightarrow k}$, whereas, in the latter scenario, pixels can belong to any set of classes, even the future ones.

Initially, a model $M_0 = D_0 \circ E_0: \mathcal{X} \mapsto \mathbb{R}^{H \times W \times |\mathcal{C}_0|}$ is trained on $\mathcal{C}_0$. 
At each subsequent step $k$, the previous step model $M_{k-1}$ and the training set $\mathcal{T}_k$ are used to develop a new model $M_k = D_k \circ E_0 : \mathcal{X} \mapsto \mathbb{R}^{H \times W \times |\mathcal{C}_{0 \rightarrow k}|}$, 
that performs well on both old ($\mathcal{C}_{0\rightarrow k-1}$) and new classes ($\mathcal{C}_k$). Notice that in our implementation the encoder $E_0$ is fixed and only the decoder is trained during the incremental steps as done in previous works \cite{michieli2019incremental}.

\section{General Architecture}
\label{sec:gen_arc}


\begin{figure*}[h]
\includegraphics[width=\linewidth]{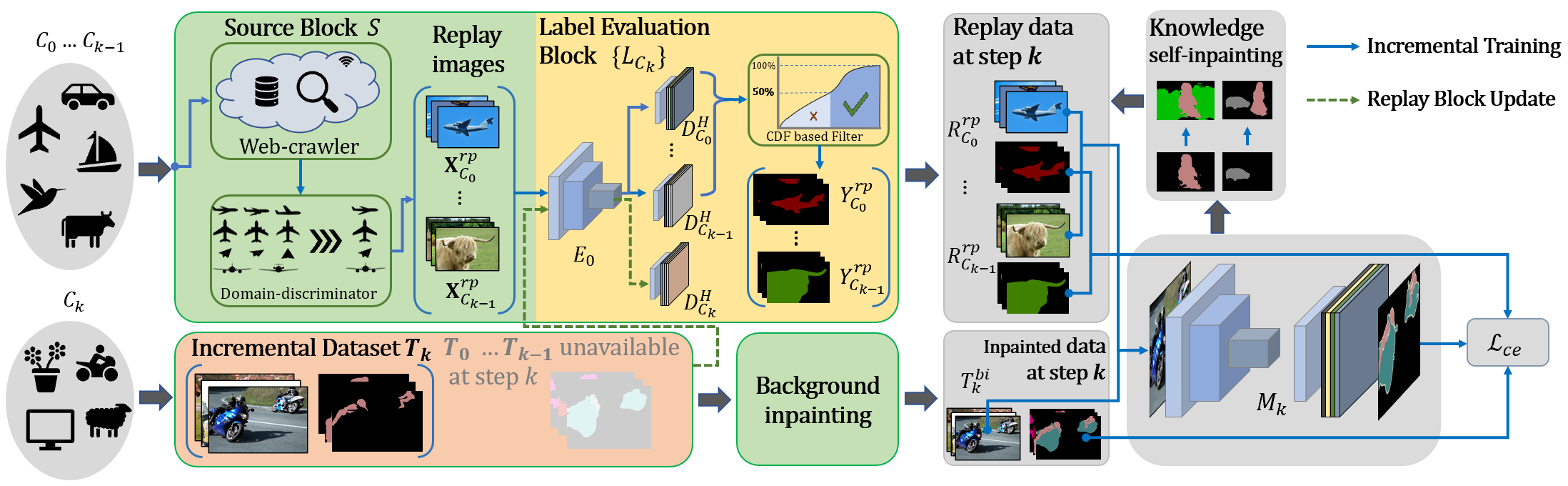}
\vspace{-0.2cm}
\caption{Overview of  RECALL+: Class labels from the past steps are retrieved by Source Block, which consists of a domain-discriminator and filters the duplicate and near-duplicate images for each class. Then these selected images are further filtered by a CDF-based thresholding strategy. Finally, the segmentation network is incrementally trained with both replay data and new class data.}


\label{fig:architecture}
\end{figure*}

%
%
%
In the incremental learning setting, when performing an incremental training step $k$\um{,} only samples related to new classes $\mathcal{C}_k$ are assumed to be available. 
Following the simplest approach, we could initialize our model's weights from the previous step ($M_{k-1}$, $k \geq 1$) and learn the segmentation task over classes $\mathcal{C}_{0 \rightarrow k}$ by optimizing the standard objective $\mathcal{L}_{ce}(M_k; \mathcal{C}_{0 \rightarrow k}, \mathcal{T}_k)$ with data from the current training partition $\mathcal{T}_k$. 
However, simple fine-tuning leads to catastrophic forgetting, being unable to preserve previous knowledge. 
%

\noindent
\subsection{Architecture of \um{the} Replay Block.}
\label{subsec:archi_replay}
To tackle this issue, a web-based replay strategy is proposed. 
Our goal is to retrieve task-related knowledge of past classes to be blended into the ongoing incremental step, all without accessing training data of previous iterations. 
To this end, we introduce a Replay Block, whose target is twofold.
Firstly, it has to provide images resembling instances of classes from previous steps, 
by retrieving them from an available 
source (\eg, a web database). 
Secondly, it has to obtain reliable semantic labels of those images, by resorting to learned knowledge from past steps. 
The Replay Block's image retrieval task is executed by what we call Source Block:
\vspace{-0.2cm}
\begin{equation}
S : \: \mathcal{C}_k \mapsto \mathcal{X}^{rp}_{\mathcal{C}_k}
\vspace{-0.2cm}
\end{equation}
This module takes in input a set of classes $\mathcal{C}_k$ (background excluded) and provides images 
whose semantic content can be ascribed to those categories 
(\eg, $\mathbf{X}^{rp} \in \mathcal{X}^{rp}_{\mathcal{C}_k}$). 
In this work, we focus on the usage of web-crawled data for this block, 
as detailed in Sec.~\ref{sec:replay}. 
%

The Source Block provides unlabeled image data (if we exclude the weak image-level classification labels), and for this reason, we introduce an additional 
Label Evaluation Block $\{ L_{\mathcal{C}_k} \}_{\mathcal{C}_k \subset \mathcal{C}}$, 
which aims at annotating examples provided by the replay module.  
This block is made of separate instances $L_{\mathcal{C}_k}$, each denoting a segmentation model to classify a specific set of semantic categories $\mathcal{C}_k \cup \{b\}$ (\ie, the classes in $\mathcal{C}_k$ plus the background):
\vspace{-0.15cm}
\begin{equation}
L_{\mathcal{C}_k} : \mathcal{X}_{\mathcal{C}_k} \mapsto \mathbb{R}^{H \times W \times (\notation{|{\mathcal{C}_k} \cup \{b\}}|)}
\vspace{-0.2cm}
\end{equation}
All $L_{\mathcal{C}_k}= D_{\mathcal{C}_k}^H \circ E_0$ modules share the encoder section $E_0$ from the initial training step so that only a minimal portion of the segmentation network $D_{\mathcal{C}_k}^H$ (\textit{helper decoders}) is stored for each block's instance (see \cite{maracani2021recall} for more details).
Notice that a single instance recognizing all classes could be used, leading to an even more compact representation, but it experimentally led to lower performance \cite{maracani2021recall}.
%

Provided that $S$ and $L_{\mathcal{C}_k}$ are available, replay training data can be collected for classes in $\mathcal{C}_k$. 
A query to $S$ outputs a generic image example $\mathbf{X}^{rp}_{\mathcal{C}_k} = S (\mathcal{C}_k) $, 
which is then associated with its prediction 
$\mathbf{Y}^{rp}_{\mathcal{C}_k}[h,w] = \argmax \limits_{c \in \mathcal{C}_k \cup \{b\}}  L_{\mathcal{C}_k}(\mathbf{X}^{rp}_{\mathcal{C}_k})[h,w][ c ] $, where $[h,w]$ indicates the pixel position.
By retrieving multiple replay examples,  a replay dataset $\mathcal{R}_{\mathcal{C}_k} = \{ ( \mathbf{X}^{rp}_{\mathcal{C}_k}, \mathbf{Y}^{rp}_{\mathcal{C}_k} )_n \}_{n=1}^{N_{r}} $ is built. Its size $N_{r}$ is an hyperparameter empirically set (see \um{S}ection \ref{sec:implementation}). 
%

\begin{figure}[tbp]
\centering
\vspace{-0.2cm}
\includegraphics[trim=6cm 3cm 5.5cm 2.5cm, clip, width=1\linewidth]{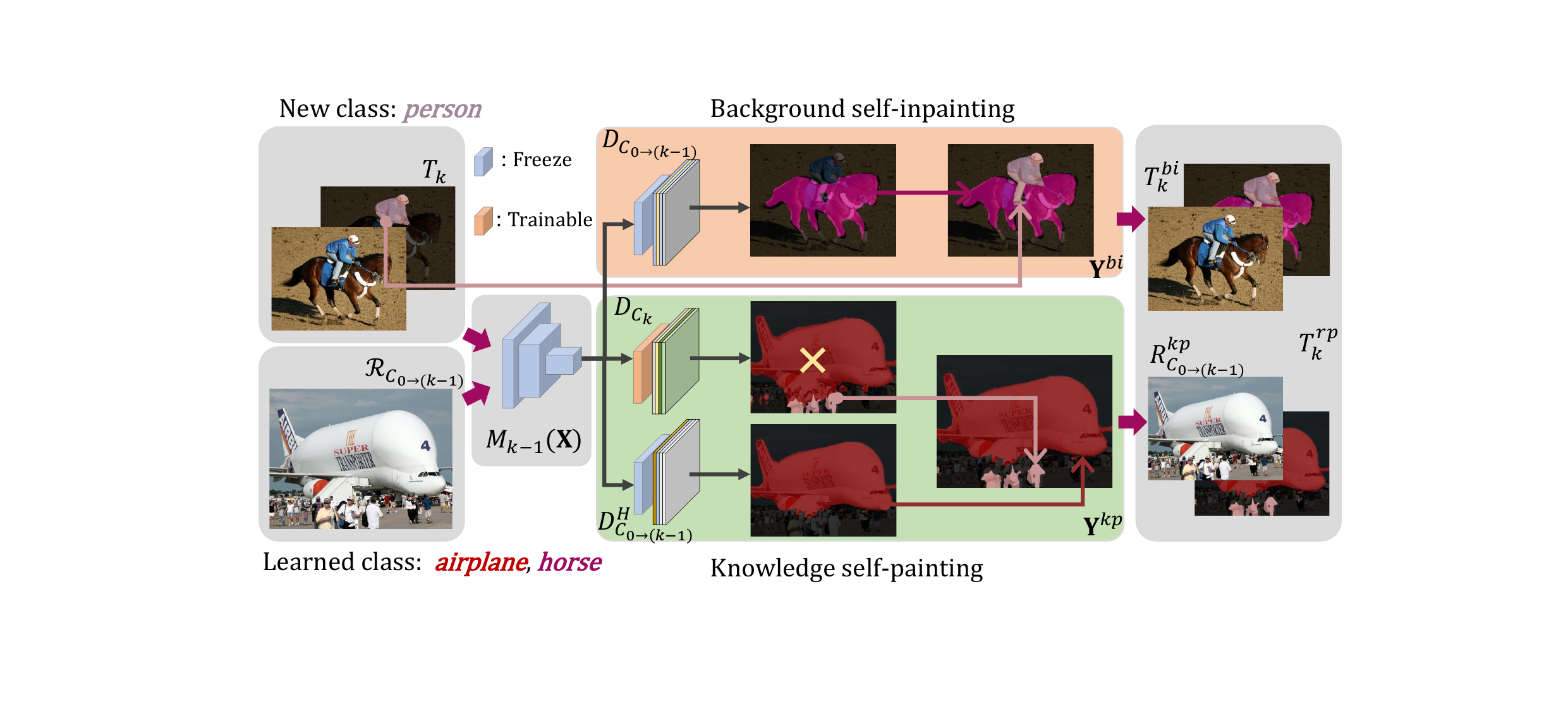}
\vspace{-0.2cm}
\caption{Background self-inpainting process and knowledge painting process. At step $\textit{k}$, the background self-inpainting technique updates the past knowledge (\ie\um{,} \textbf{\textit{\color{RedViolet}horse}}) on the current step training images before the training step starts. During training, the knowledge \um{self-in}painting updates the label information of the classes being learned (\ie, \textbf{\textit{\color{Salmon}person}}) to the web downloaded images.} 
\label{fig:bi_and_bp}
\vspace{-0.1cm}
\end{figure}


\noindent
\subsection{Self-teaching Strategies.}
\label{sec:ST}
The background shift phenomenon is a common challenge in continual learning for semantic segmentation. It refers to the fact that the background class distribution changes over time during the learning process. As new classes (whose pixels in previous steps were labeled as background) are introduced, the context and appearance of the background class may vary.  

To deal with the background shift phenomenon, we propose a novel inpainting mechanism to introduce knowledge from the previous and current models into both the current step images and the web-crawled data. 
While in the conference version \um{\cite{maracani2021recall}}, background inpainting was used only on background regions of current samples, here we use it also on web-crawled data.

\textbf{Background Self-Inpainting.}
The first strategy, derived from the conference version, aims at reducing the background shift and\um{,} at the same time\um{, at} bringing a regularization effect similar to knowledge distillation \cite{michieli2019incremental,cermelli2020modeling}. 
Despite serving the same purpose, it differs in the implementation.

At every step $k$, we take the background region of each ground truth map of the training set $\mathcal{T}_k$ and we label it with the associated prediction from the model at the previous step $M_{k-1}$ (see Fig.~\ref{fig:bi_and_bp}). We call this procedure \textit{background inpainting} since the background regions in label maps are updated according to a self-teaching scheme based on the prediction of the old model.
More formally, we replace each original label map $\mathbf{Y}$ available at step $k>0$ with its inpainted version $\mathbf{Y}^{bi}$: 
\vspace{-0.1cm}
\begin{equation}
\mathbf{Y}^{bi} [h,\!w]  \!=\!                                                                                                                                                                                   
\begin{cases}
    \mathbf{Y} [h,\!w]           &  \text{if} \: \mathbf{Y} [h,\!w]\! \in\! \mathcal{C}_k \\
     \argmax \limits_{c \in \mathcal{C}_{0 \rightarrow k-1}}  M_{k - 1} (\mathbf{X})   [h,\!w][c]  &  \text{otherwise}
\end{cases}
\end{equation}
where $(\mathbf{X},\mathbf{Y}) \in \mathcal{T}_{k}$, while $[h, w]$ are the pixel coordinates. Labels at step $k=0$ are not inpainted, as in that case there are no previous step classes. 
Specifically, we utilize pseudo-labels generated from the previous model to fill in the background areas that exist in the ground truth label.
When background inpainting is performed, each set $\mathcal{T}_{k}^{bi} \subset \mathcal{X}\times \mathcal{Y}_{\mathcal{C}_{0 \rightarrow k}}$ ($k>0$) contains all samples of $\mathcal{T}_k$ after being inpainted.

\noindent
\textbf{Inpainting \um{W}eb \um{D}ata with \um{C}urrent \um{C}lasses' \um{K}nowledge.}
Images downloaded from the web typically contain also instances of the set of classes currently being learned.
As an example, in images of the \textit{bicycle} or \textit{sofa} classes, it is likely that there \um{are also} instances of the \textit{person} class.
The helper decoders can only label already learned classes: \um{\eg, if} we are learning the \textit{person} class, the optimization for the corresponding instances existing in the replay data are labeled as background during the training.
To exploit also this information, we let the main model update the labeling of the replay images during training, namely \textit{knowledge self-inpainting} (see Fig. \ref{fig:bi_and_bp}). 
Notice that, in the case of knowledge self-inpainting, the model is already familiar with the information from the replayed classes. This introduces the possibility of expanding the old-class pixels through self-inpainting, which could mislead the model and lead to a drop in performance. To address this concern, we introduce a third term in the process of knowledge inpainting. This term acts as a constraint, ensuring that the background pixel labels predicted by the current model as old classes remain unchanged. This constraint prevents the model from being misled by self-inpainting, thereby maintaining performance stability (see Section \ref{subsec:abl_inpaint}).

The knowledge self-inpainting strategy can be defined as:  
\vspace{-0.1cm}
\begin{equation}
\label{eq:knowledge_inpainting}
\mathbf{Y}^{ki} [h,\!w]  \!=\!  
\begin{cases}
    \mathbf{Y^{rp}} [h,\!w]           &  \text{if} \: \mathbf{Y^{rp}} [h,\!w]\! \in\! \mathcal{C}_{\mathcal{C}_{0 \rightarrow k-1}} \\
    \mathbf{Y^{max}} [h,\!w]           &  \text{if} \: \mathbf{Y^{rp}} [h,\!w]\! \notin\! \mathcal{C}_{\mathcal{C}_{0 \rightarrow k-1}} \\
    & \land \mathbf{Y^{max}} [h,\!w] \in \mathcal{C}_k \\
    b & \text{otherwise}
\end{cases}
\end{equation}
Notice how the labels are updated only in the second case of Equation (\ref{eq:knowledge_inpainting}), i.e., for background pixels that are labeled with one of the currently being learned classes.

%
%
\noindent
\subsection{Incremental Training with Replay Block.}
The training procedure of the proposed approach is summarized in Algorithm\ \ref{algorithm} and depicted in Fig.~\ref{fig:architecture}. 
Let us focus on a generic incremental step $k$, where only samples of classes in $\mathcal{C}_k$ from partition $\mathcal{T}_k$ are available. 
%
Firstly, the Replay Block is used to retrieve unlabeled web data for steps from $0$ to $k-1$ among all the past classes.
Then, the domain discriminator followed by the semantic selection technique is applied over the replay images of each incremental class set  $ \mathcal{C}_i, i=0,...,k-1$. 
The replay training dataset for step $k$ is the union of the  replay sets of all previous steps: $\mathcal{R}_{\mathcal{C}_{0 \rightarrow (k-1)}} = \bigcup\limits_{i=0}^{k-1} 
\mathcal{R}_{\mathcal{C}_i}$. 
Once we have assembled $\mathcal{R}_{\mathcal{C}_{0 \rightarrow (k-1)}}$, by merging it with $\mathcal{T}_k^{bi}$ we get an augmented step-$k$ training dataset $\mathcal{T}_k^{rp} = \mathcal{T}_k^{bi} \cup \mathcal{R}_{\mathcal{C}_{0 \rightarrow (k-1)}}$.   
This new set, in principle, includes annotated samples containing both old and new classes, thanks to replay data. 
Therefore, we can effectively learn the segmentation model $M_k$ through the cross-entropy objective {$\mathcal{L}_{ce}(M_k; \mathcal{C}_{0 \rightarrow k}, \mathcal{T}_k^{rp})$} on replay-augmented training data.
This mitigates the bias toward new classes, thus
preventing catastrophic 
 forgetting. 
%
%

Then, we fine-tune the domain discriminator using the $\mathcal{T}_k \cup \mathcal{R}_{\mathcal{C}_{0 \rightarrow (k-1)}}$ set (see Sec.~\ref{sec:AL}) and optimize the decoder $D^H_{\mathcal{C}_k}$ 
to segment images from $\mathcal{T}_k$ by minimizing $\mathcal{L}_{ce}(L_{\mathcal{C}_k}; \mathcal{C}_k \cup \{b\}, \mathcal{T}_k)$. These stages are not necessary for the current step, but will be exploited for the future ones. 

During a standard incremental training stage, 
we follow a mini-batch gradient descent scheme, where batches of annotated training data are sampled from $\mathcal{T}_k^{rp}$.
However, to guarantee a proper stream of information, we opt for an interleaving sampling policy, rather than a random one. 
In particular, at a generic iteration of training, a batch of data $\mathcal{B}^{rp}$ supplied to the network is made of $r_{new}$ samples from the current training partition $\mathcal{T}_k^{bi}$ and $r_{old}$ replay samples from $\mathcal{R}_{\mathcal{C}_{0 \rightarrow (k-1)}}$. 
The ratio between $r_{new}$ and $r_{old}$ controls the proportion of replay and new data (details are discussed in \cite{maracani2021recall}). 
We need, in fact, to carefully \um{control} how new data is balanced with respect to replay one so that enough information about new classes is provided within the learning process, 
while concurrently we assist the network in recalling knowledge acquired in past steps to prevent catastrophic forgetting. 

\section{Image Selection Strategies}
\label{sec:replay}

We assume to collect images from a generic photo-sharing website.
\um{For the results we use Flickr}, which proved to be a good choice for multiple reasons. 1) 
\um{Images from Flickr
contain user-uploaded pictures} that implicitly contain real world noise, making the model more robust during deployment; 2) differently from machine learning training datasets, in which 
\um{images may be acquired in a controlled setup,}
these images are collected in-the-wild, thus covering a wide range of pose and lighting conditions, making the dataset \um{more} unbiased and
diverse, but also less reliable; 3) using randomly selected uploaded images, additionally, allows to reduce the domain gap between the training data and real-world deployment.
Notice that the usage of Flickr images requires to adhere to the corresponding terms of use, however the approach is agnostic to the web data source, if licensing is a concern one can select images having the desired license (e.g., Creative Commons CC-0), indeed any suitable repository can be chosen. 

\um{Querying} for photos on the web \um{via} the class label name 
yields uncontrolled results that include images useful for the training, but also images without the expected content or with other 
\um{anomalies} (see Fig. \ref{fig:diverse} and Fig. \ref{fig:discriminator-fail})
making them \um{useless} or even misleading. 
In this section, we present the  strategies \um{we introduce} in this extended version
to select suitable pictures from the web data for training.
To this aim, we use the combination of an adversarial learning approach with a threshold-based selection mechanism. Firstly, we use the adversarial learning strategy to filter out images with \um{statistical distributions} that do not resemble the training dataset ones. 
Then, an adaptive threshold filtering strategy based on the predicted segmentation chooses only the images that have enough pixels predicted as  belonging to the selected class. 

\subsection{Adversarial Learning for Image Selection. }
\label{sec:AL}
It is crucial for a training dataset to be unbiased, to have enough diversity and to properly capture the statistical properties of the different classes. 

To get the desired unbiased \fb[]{and diverse} images, we 
downloaded \fb[]{them} from the Flickr website since it contains billion\um{s of}
images uploaded by common users. 
Then, to ensure diversity, we 
\um{discarded} duplicate images 
by a simple thresholding based on the peak signal-to-noise ratio (PSNR) between couples of images.



Following the download and verification process, we obtained a collection of 10K \fb[]{unique} images for each class\um{. We remark} that a practical system should download images on-the-fly during training, however we downloaded a fixed set to ensure repeatability of results across multiple experiments. \fb[]{Yet},  
this does not guarantee that these images have statistical properties suitable for the training.
We assume that the images in the Pascal dataset used for supervised training had the desired characteristics, and we aim to find web images that closely resemble the no longer available Pascal ones.

\textbf{Adversarial Training strategy.} 
For this purpose, we train a discriminator network to distinguish between Pascal images and web-downloaded ones. 
Rather than training an individual discriminator network for each incremental step, which would lead to a memory consumption increase proportional to the number of steps, 
we utilize a single discriminator and employ the following incremental training approach:

(1) at step $0$\um{,} the negative samples are selected from the web images for classes in $\mathcal{C}_{0}$ and the positive samples are taken from the Pascal data also for classes  in $\mathcal{C}_{0}$; 

(2) for each incremental step $k>0$, the images of the currently learned classes are selected using the same strategy, however, for the no-longer available positive samples of the old classes, we used web-images that prove effective at deceiving the discriminator. 
 These images do not increase the memory burden since they can be easily obtained through web queries at each step $k$.

More in detail, given an input $\mathbf{X}$, the discriminator produces confidence score (logits) $z = [z_p, z_{rp}] \in \mathbb{R}^2_{0+}$ for training dataset images ($z_p$) and web-replay ones ($z_{rp}$), where negative values are excluded since $z$ is passed through a ReLU activation function.
Normally, the discriminator compares $z_p$ and $z_{rp}$ to produce a classification outcome: if $z_p > z_{rp}$, $\mathbf{X}$ is classified as original training data, and vice versa.
However, this relationship is not strong enough to select the images that will be used to update the discriminator, therefore we introduce the set of \textit{core} Replay Images $\mathcal{R}_{\mathcal{C}_{0\rightarrow (k-1)}}^{core}$ defined as follows:

\begin{equation}
    \mathcal{R}_{\mathcal{C}_{0\rightarrow (k-1)}}^{core} = \{ \mathbf{X}^{rp}_{\mathcal{C}_{0\rightarrow (k-1)}} \: | \: z_{p} > \alpha \: z_{rp} \}
\end{equation}
Here the hyper-parameter $\alpha$ controls the rate between the two scores and we set it much larger than $1$ (for the results we empirically set $\alpha = 100$). This means that only the replay samples that mislead the discriminator with very high confidence are chosen in order to achieve a more accurate training for the old classes.  

\begin{figure}[tbp]
\centering
\vspace{-0.2cm}
\includegraphics[trim=2.5cm 10cm 7.5cm 2cm, clip, width=1\linewidth]{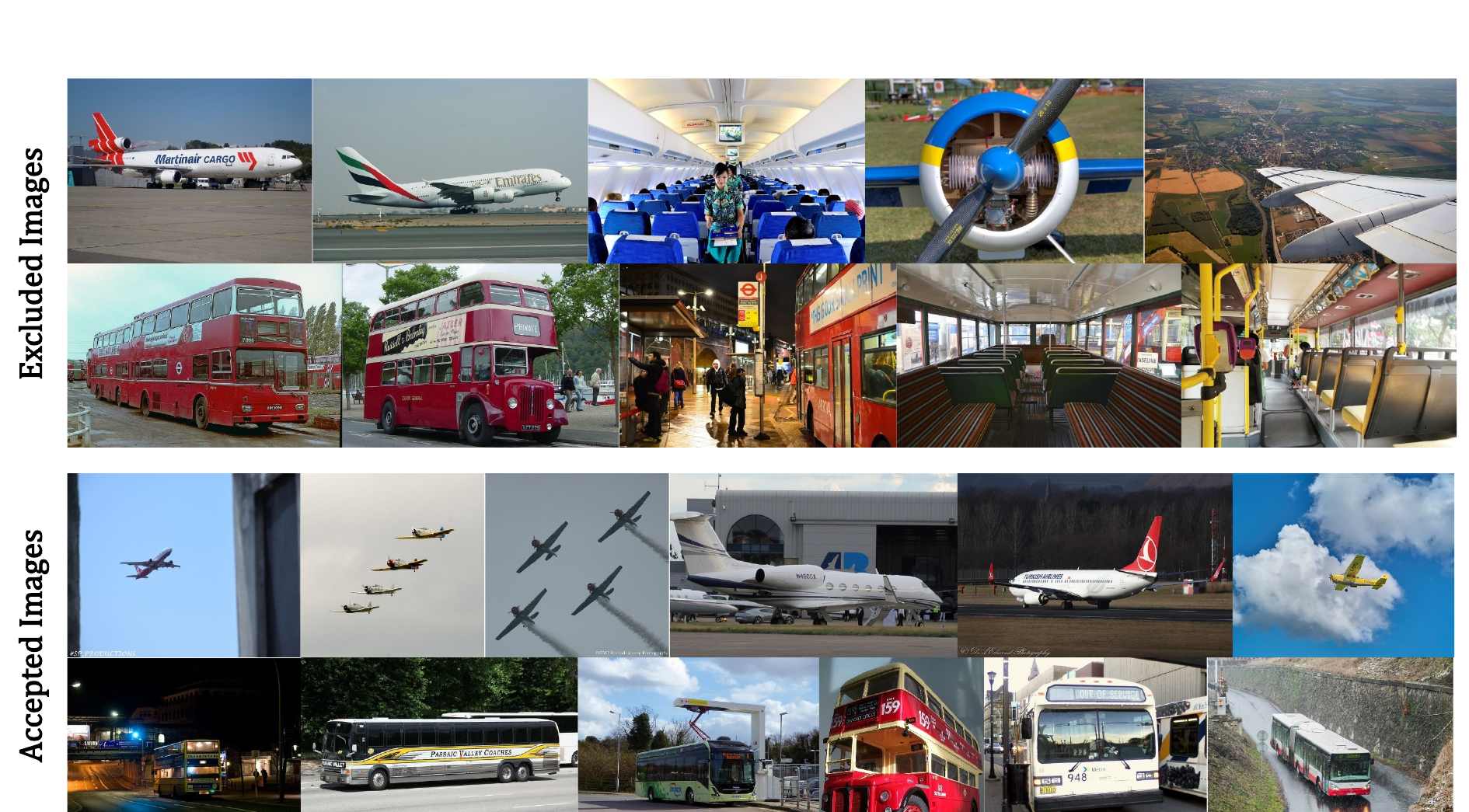}
\vspace{-0.2cm}
\caption{Examples of filtered and accepted images according to our adversarial image selection strategy. 
} 
\label{fig:diverse}
\vspace{-0.1cm}
\end{figure}

To summarize, in each incremental step $k$ the discriminator is trained using $\mathbf{X}^{rp}_{\mathcal{C}_{0\rightarrow k}}$ as negative samples, and $\mathbf{X}_{\mathcal{C}_k} \bigcup \mathcal{R}_{\mathcal{C}_{0\rightarrow (k-1)}}^{core}$ as positive samples. Notice that, $\mathbf{X}^{rp}_{\mathcal{C}_{0\rightarrow (k-1)}} \: \bigcap \: \mathcal{R}_{\mathcal{C}_{0\rightarrow (k-1)}}^{core} = \varnothing$. 
The network and implementation procedure are detailed in Section \ref{sec:implementation}.
The discriminator is intentionally trained to achieve a reasonably good accuracy, but not excessively high. This allows us to utilize it as a filter: when we obtain a new image from the web, we input it to the discriminator and retain only the images that successfully deceive the discriminator and are classified as original training set images. This approach enables us to preserve the images that show properties closely resembling those of the original training set.
Some examples of the selected and discarded images are shown in Fig.~\ref{fig:diverse}. It can be observed that the discriminator-selected images exhibit a larger diversity.


\subsection{Image Selection with Semantic Content}

\begin{figure}[tbp]
\includegraphics[width=1\linewidth]{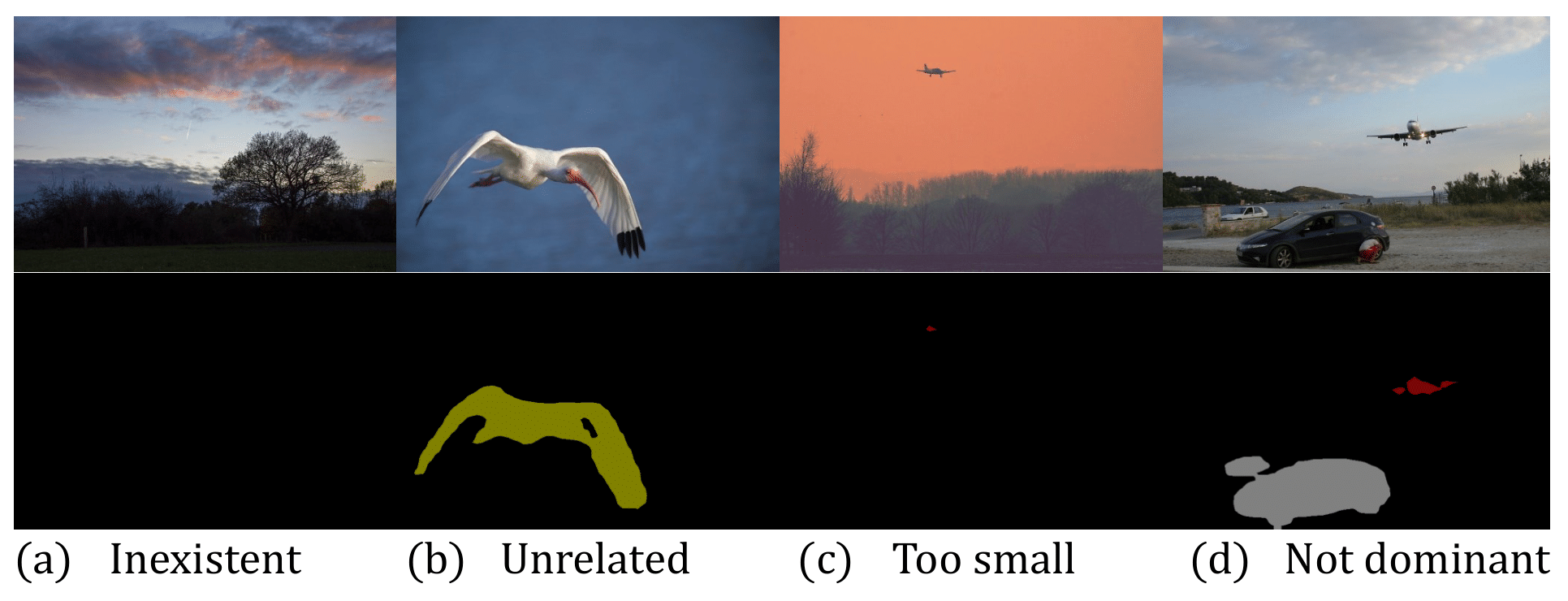}
\caption{Examples of airplane images 
that provide erroneous input from a semantic point of view. We can recognize four cases: missing class, wrong class, object too small, and non-dominant class. These cases are tackled by CDF-based filtering.}
\label{fig:discriminator-fail}
\end{figure}

Despite reducing the domain shift between replay and training images, the domain discriminator is not able to distinguish between a meaningful sample and a less relevant one (both in the semantic and spatial sense). As visible in Fig.~\ref{fig:discriminator-fail}, some web-selected \textit{airplane} images  
show a lack of information to a human observer.
To filter out non-helpful images, we rely on the currently trained segmentation network to further analyze their information content. 
The rationale is to preserve only the images that have a reasonable number of pixels belonging to the expected class.

\begin{figure}[tbp]
\centering
\vspace{-0.2cm}
\includegraphics[trim=2.5cm 12cm 3cm 2.5cm, clip, width=1\linewidth]{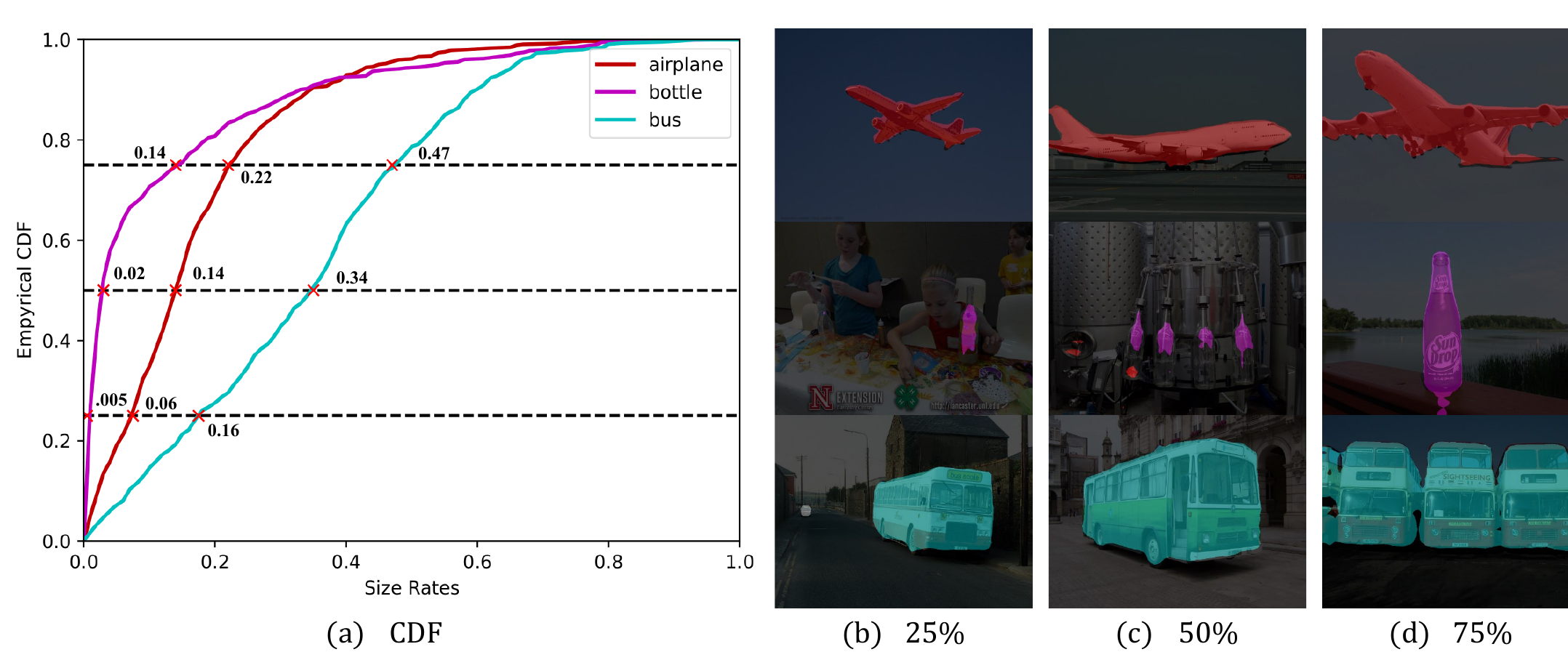}
\vspace{-0.2cm}
\caption{Visualization of the pixels with different ratios. The CDF curves of different classes differ from each other.
} 
\label{fig:cdf}
\vspace{-0.1cm}
\end{figure}

Different objects have different sizes and using a fixed threshold for all classes proved to lead to unsatisfactory performance. 
To tackle this issue, we analyzed the  probability distribution functions of the fraction of pixels for each class into the corresponding images to get a reference for thresholding. 
That is, for each class $c$ we computed the Cumulative Distribution Function (CDF, $\mathcal{F}_c$) of the  distribution $\mathcal{P}_{(\mathbf{X},\mathbf{Y}) \sim \mathcal{T}_{k} | c \in \mathcal{C}_{0\rightarrow k}} ^c[\mathbf{Y} = c]$ and we used it to extract suitable thresholds for the object sizes according to a quantile strategy.
Therefore we computed the threshold value $t^{size}_{c}$ for the number of pixels of class $c$ in an image correspoding to that class  as follows:
\begin{equation}
    t^{size}_c = \mathcal{F}_c^{-1}(0.5)
\end{equation}
The samples considered acceptable by our strategy are those whose fraction of pixels in the corresponding class belongs to $[t^{size}_c, 1]$ ($t^{size}_c$ equals to the median of distributions). 
An example is reported in Fig. \ref{fig:cdf} (a): the plot shows the CDF curves for three representative classes. In around $80 \%$ of the  \textit{bottle}  images the object pixels  are no more than $20 \%$ of the total  image size. On the other side over half of the \textit{bus} images have buses with a relative size of more than $30 \%$, which shows how there is a large difference from one class to another. Fig. \ref{fig:cdf} (b)-(d) shows some examples of the pixels of three sample classes at different thresholds. 



\begin{algorithm}[tbp]
\caption{RECALL+: incremental training procedure.}
\label{algorithm}
\vspace{1mm}
\textbf{Input:} $\{ \mathcal{T}_k \}_{k=0}^K$ (Sec. \ref{sec:problem}) and $\{ \mathcal{C}_k \}_{k=0}^K$ (Sec. \ref{sec:problem})
\\
\textbf{Output:} $M_K$ (Sec.\ref{sec:problem}) \\[0.3ex]
train $M_0 = E_0$  $\circ D_0$  with $\mathcal{L}_{ce}(M_0; \mathcal{C}_{0}, \mathcal{T}_0)$ (Sec. \ref{sec:gen_arc}) \\
train $D^H_{\mathcal{C}_0}$ (Sec.\ref{subsec:archi_replay}) with $\mathcal{L}_{ce}(L_{\mathcal{C}_0}$ $; \mathcal{C}_{0}, \mathcal{T}_0)$ \\
\textbf{for} $k \leftarrow 1$ to $K$ \textbf{do} \\
\hspace*{4mm} background inpainting on $\mathcal{T}_k$ to obtain $\mathcal{T}_k^{bi}$ \\
\hspace*{4mm} train $D^H_{\mathcal{C}_k}$ with $\mathcal{L}_{ce}(L_{\mathcal{C}_k}; \mathcal{C}_{k} \cup \{ b \}, \mathcal{T}_k)$ \\
\hspace*{4mm} get candidate images $\mathcal{R}_{\mathcal{C}_{0 \rightarrow (k-1)}}$ (Sec. \ref{subsec:archi_replay}) from the web  \\
\hspace*{4mm} perform \textit{adversarial learning} and \textit{semantic content} selection on $\mathcal{R}_{\mathcal{C}_{0 \rightarrow (k-1)}}$  \\
\hspace*{4mm} generate $\mathcal{T}_k^{rp} = \mathcal{T}_k^{bi} \cup \mathcal{R}_{\mathcal{C}_{0 \rightarrow (k-1)}}$ \\
\hspace*{4mm} train $D_k$ with $\mathcal{L}_{ce}(M_k; \mathcal{C}_{0 \rightarrow k}, \mathcal{T}^{rp}_k)$ \\
\hspace*{4mm} knowledge inpainting on $\mathcal{R}_{\mathcal{C}_{0 \rightarrow (k-1)}}$ \\
\hspace*{4mm} continual training $D_k$ with $\mathcal{L}_{ce}(M_k; \mathcal{C}_{0 \rightarrow k}, \mathcal{T}^{rp}_k)$ \\
\hspace*{4mm} fine tune discriminator on $\mathcal{T}_k\cup\mathcal{R}_{0\rightarrow k}$ \\
\textbf{end for}
\vspace{1mm}
\end{algorithm}




\section{Implementation Details}
\label{sec:implementation}
Following other works in continual semantic segmentation, we use DeepLab-V3 \cite{chen2017rethinking} with ResNet-101 \cite{he2016deep} as our base architecture. Nonetheless, notice how the proposed strategy is independent of the backbone architecture. 
The encoder has been initialized using a pretrained model on ImageNet \cite{deng2009imagenet} and all the parameters are trained during the initial step $0$.
For the following incremental steps, we freeze the encoder part thus only the main decoder is trained, together with additional $\{ D^{H}_{\mathcal{C}_{k}} \}_k$ helper decoders, which are used for annotating replay samples in  future steps.
For fair comparison, we used the same training parameters of the conference version \cite{maracani2021recall} for the initial model $M_0$.
Following \cite{michieli2021continual}, we crop images to $512 \times 512$  and apply  data augmentation (\ie, random scaling the images of a factor from 0.5 to 2.0 and random left-right flipping). 
We adopt SGD for weight optimization, with initial learning rate $5\times10^{-4}$ decreased to $1\times10^{-4}$ by  polynomial decay of power $0.9$. 
At each step we train the model for $|\mathcal{C}_k| \times 1000$  steps  in both disjoint and overlapped setups (e.g., around 4 epochs in task 10-5-5). 
The knowledge self-inpainting is performed once when training progress reaches around 60\% of total training steps.
Each helper decoder $D^{H}_{C_{k}}$ is trained with a polynomial decaying learning rate starting from $2\times10^{-4}$ and ending at $2\times10^{-6}$ for $|\mathcal{C}_k| \times 1000$ steps.
The interleaving ratio $r_{old}/r_{new}$ is set to $1$.
For the web images we downloaded a set of $10000$  images for each class (this allowed to perform all the tests with the same data thus ensuring reproducibility) and apply to them the selection strategies of Section \ref{sec:replay}.
For the experiments on Pascal VOC 2012 \cite{everingham2012pascal}, we exploit the first 500 selected replay images of each class for training  (the first 100 replay images for ADE20K \cite{zhou2017scene}).

The discriminator of the adversarial module consists of a  pre-trained EfficientNet-B0  encoder \cite{tan2019efficientnet} followed by a three-layer fully connected network (FCN) with dimensions $1000$, $256$ and $2$. Training images are resized to $224 \times 224$. We employed the dataset images  used for the training  as positive samples and web-downloaded images as negative samples. A chunk of $20\%$ of the training set is let out for  validation. 
The discriminator is trained for 10 epochs using SGD with an initial learning rate set to $1\times10^{-3}$ and decreased with a rate of $0.8$ every 2 epochs. 
The confidence parameter $\alpha$ is set to 100 during the discriminator training.
The entire framework is implemented with Pytorch \cite{paszke2019pytorch} and trained on a single NVIDIA GTX 1080 Ti. Training time varies depending on the setup, with the longest run taking about $10$ hours. 

\input{sections/results_tables_1_2}

\section{Experimental Results}
\label{sec:results}


In this section, we present the experimental evaluation on the Pascal VOC 2012 \cite{everingham2012pascal} and ADE20K \cite{zhou2017scene} datasets. 
Following previous works  \cite{shmelkov2017incremental,michieli2019incremental,michieli2021knowledge,cermelli2020modeling}, we start by analyzing the performance on 3 widely used incremental scenarios: \ie, the addition of the last class (19-1),  the addition of the last 5 classes at once (15-5) and the addition of the last 5 classes sequentially (15-1). 
Moreover, we report the performance on three more challenging scenarios in which 10 classes are added sequentially one by one (10-1), in 2 batches of 5 elements each (10-5) and all at once (10-10).
Classes for the incremental steps are selected according to the alphabetical order. 
 The naive fine-tuning approach (FT) represents the lower limit to the accuracy of an incrementally learned model, while the joint training on the complete dataset in one step corresponds to the upper bound. We also report the results 
of a simple Store and Replay (S\&R) method, where at each incremental step we store a certain number of true samples for newly added classes, such that the respective size  in average matches the size of the helper decoders needed by RECALL+. 
As comparison, we include $2$ methods extended from classification (\ie, LwF \cite{li2018learning} and its single-headed version LwF-MC \cite{rebuffi2017icarl}) 
and some of the most relevant methods 
designed for continual segmentation (\ie, ILT \cite{michieli2019incremental}, CIL \cite{klingner2020class}, MiB \cite{cermelli2020modeling}, SDR \cite{michieli2021continual}, RECALL\cite{maracani2021recall},  PLOP\cite{douillard2021plop}, RCIL\cite{zhang2022representation}, REMI\cite{phan2022class}, SPPA \cite{lin2022continual}, RBC\cite{zhao2022rbc}, AWT\cite{goswami2023attribution} and EWF\cite{xiao2023endpoints}). 
%
%
%
Exhaustive quantitative results in terms of mIoU 
are shown in Table \ref{tab:voc_1} for Pascal dataset and Table \ref{tab:ade} for ADE20K. 
For each setup we report the mean accuracy for the initial set of classes, for the classes in the incremental steps and for all classes, computed after the overall training.
Besides, since the performance of the joint training is different, here we give another metric namely ``$\Delta$'' (the smaller the better) to evaluate the proposed approach. More in detail, $\Delta$ is given by the difference between the joint and the incremental models (respectively {Joint} and \textit{all}, in both Table~\ref{tab:voc_1} and ~\ref{tab:ade}).

\noindent
\textbf{Addition of the last class.} 
In the first basic experiment, the model is  trained over the first 19 classes during the initial step and then a single incremental step 
on the  \textit{tv/monitor} class is performed. 
Looking at 
Table~\ref{tab:voc_1}, fine-tuning leads to a large performance drop. Using replay data from the web and inpainting strategies, our approach is able to preserve past knowledge and properly learn the new classes. Notice how there is a performance improvement of about $5\% $ in the disjoint scenario and $2\%$ in the overlapped one, showing how the new image selection strategies allowed for a much better data replay than in the previous conference version. The overall mIoU is better than most competitors even if a couple of the most recent ones outperform our approach, but as already pointed out our approach is optimized for more realistic settings with many incremental steps. 

\input{sections/table_ade}
\input{sections/results_table_delta}
\noindent
\textbf{Addition of last 5 classes.}
Then, we considered two more challenging settings where 15 classes are learned in the initial step and the last 5 in an incremental way, in one shot (15-5) or sequentially one at a time (15-1), respectively, leading to a more severe catastrophic forgetting.

%
Taking a closer look at the results in Table~\ref{tab:voc_1} (upper mid and right sections), FT has very low performances in this case and continual learning methods are required. 
Again the proposed approach effectively tackles catastrophic forgetting. 
The trend can be observed both in the disjoint and overlapped settings in both the 15-5 setup and more evidently in the 15-1 one.  
Notice how exploiting web replay samples proves to effectively restore knowledge of past classes and the additional refinements introduced in this journal version allow for a large performance increase in all settings, compared to RECALL. 

When comparing with competitors it is possible to notice how our approach scales much better when multiple incremental steps are performed: in the 15-5 setting\um{,} results are similar to the best competitors (nevertheless, according to the $\Delta$ metric, that reflects the best-case scenario, our approach stands out as the top performer). In the 15-1 (where 5 incremental steps are performed), we clearly outperform all competitors.  
%
%
%

\noindent
\textbf{Addition of last 10 classes.}
For further analysis, we consider even more challenging settings where the initial step contains only 10 classes. 
The other 10 can be learned in a single step (10-10), in 2 steps of 5 classes (10-5), or one-by-one (10-1). 
Especially the latter two settings where the learning happens in multiple stages prove to be very challenging for most continual learning approaches. 
The introduction of replay data allows for a very large performance improvement, 
outperforming competitors in the settings with more learning steps. 
In this case, the gap with the conference version  of the work \cite{maracani2021recall} is a bit smaller, but notice that in the  most challenging setting (10-1), the new approach has the best performance in both overlapped and disjoint settings with a noticeable gap with respect to the conference version and other competitors.
As a final remark, notice that on one side it is true that the approach exploits additional external data but, on the other side, note how this is unsupervised data easy to obtain from the web. Furthermore, we do not exploit any additional provision in the network training like additional modules or loss functions. 

\noindent
\textbf{ADE20K.}
Table \ref{tab:ade} shows the results on the challenging ADE20K dataset \cite{zhou2017scene}. 
The first considered setting is 100-50 (2 steps). Here the web-crawled images allow us to outperform competitors. 
The 100-10 setting has 6 steps: despite the baseline network performance being worse than some competitors, our approach still gains competitive performance, as proved by the $\Delta$ metric that is the second best. 
For the more challenging setting 100-5 (11 steps), RECALL+ achieves the state of the art (w.r.t.\ the $\Delta$ metric), outperforming other methods, which verifies the robustness of RECALL+ in tasks with a large number of incremental steps.
We also notice that for the 50-50 task, RECALL+ is slightly weak\um{er} than some competitors. We conjecture that both the frozen backbone and the large number of new classes in every single step drop the performance.

\begin{figure*}[htbp]
\centering
\vspace{-0.15cm}
\includegraphics[trim=4cm 13.8cm 14cm 3cm, clip, width=.9\linewidth]{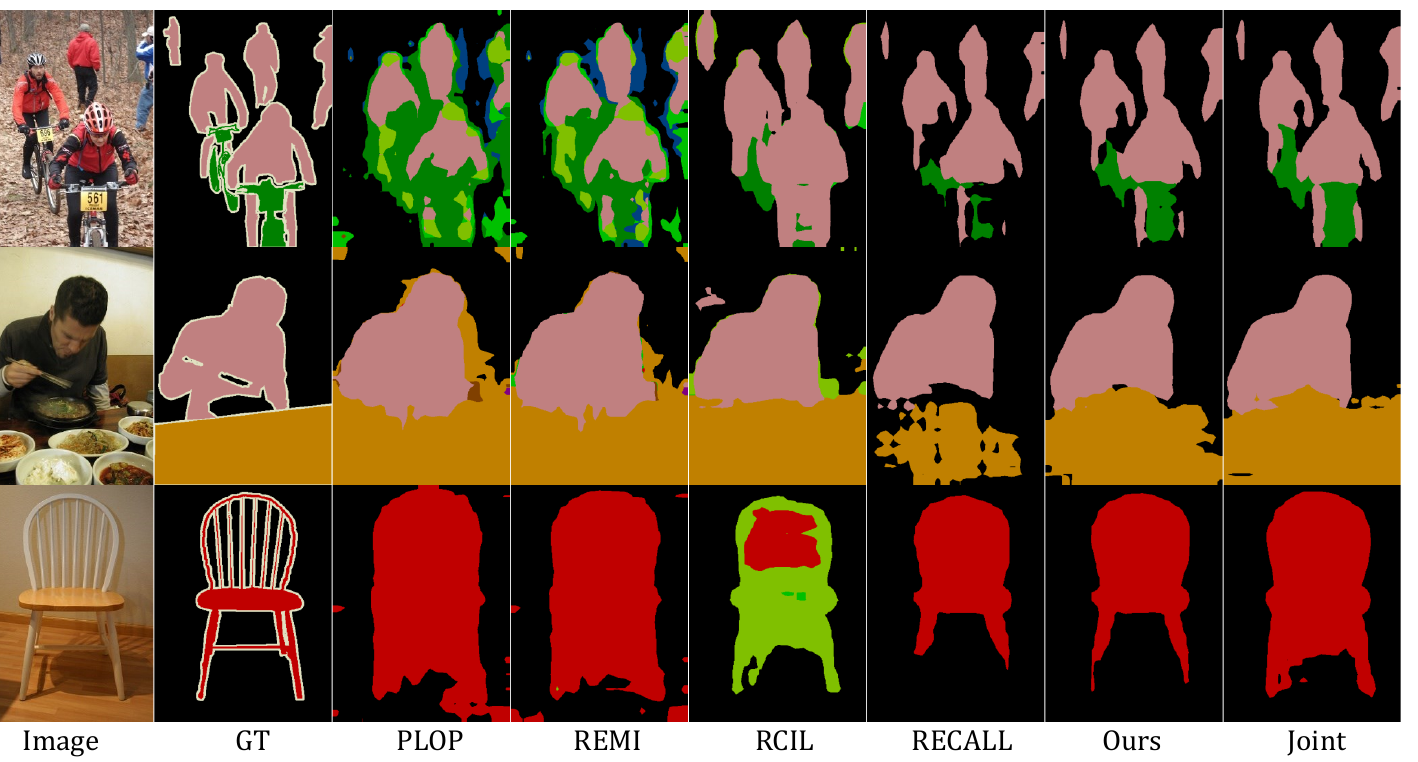}
\vspace{-0.2cm}
\caption{Qualitative results comparison on disjoint incremental setup 15-1 \um{on the} Pascal \um{VOC} 2012 validation dataset.} 
\label{fig:qualitative}
\vspace{-0.1cm}
\end{figure*}

\noindent
\textbf{Qualitative evaluation.}
We further show the qualitative results of different methods in the 15-1 disjoint setting in Fig.~\ref{fig:qualitative}.
In the first row, PLOP and REMI are confused by the background, which causes mislabeling of the \textit{person}. RCIL and RECALL preserve the knowledge of \textit{person}, while the shape of the \textit{bike} is partially lost. Compared with other methods, RECALL+ not only prevent mislabeling, but preserve most area of \textit{bike} and \textit{person} w.r.t joint-trained model predicted label.
For the second row, PLOP and REMI lose control of the \textit{dining table} and few \textit{potted plant} pixels occurred on the RCIL predicted labels. RECALL preserves the \textit{person} well but the information about \textit{dining table} is also forgotten. By using the proposed image-selection strategy, diverse replay images help our RECALL+ preserve the information about \textit{dining table} and prevent prediction to be more coarse. 
Finally, for the last row, similar to the former situation, PLOP and REMI give more coarse predictions, and RCIL totally mislabels the \textit{chair} to the \textit{potted plant}. Compared with RECALL, beneficial from the replay images and powerful helpers, the prediction by RECALL+ is more similar to the ground truth label, showing the effectiveness of the proposed image-selection strategy.

\section{Ablation Study}
\label{subsec:ablation}

To further validate the effectiveness  of the various components of our approach, we perform some ablation studies.
Firstly, we discuss the impact of the background and knowledge self-inpainting techniques introduced in Section \ref{sec:ST}. Then, we evaluate the proposed image selection methods to evaluate their impact on the achieved accuracy.

\noindent \textbf{Inpainting Strategies.}
\label{subsec:abl_inpaint}
We start by analyzing the contribution of the background self-inpainting and knowledge self-inpainting in the most challenging task, \ie, the  10-1 setting.
Results are presented in Table \ref{tab:abl_all}: 
the background inpainting technique provides a solid contribution to the preservation of previous labels, as expected, since it forces the optimization to focus also on data of the past classes, playing a role similar to knowledge distillation strategies used in other approaches.
By inpainting the previous labels in the new images, we achieve an improvement of  1.9\% (from 64.0\% to 65.9\%) and 10.9\% (from 45.4\% to 56.3\%) on old and new classes, respectively.
The knowledge self-inpainting on replay data  brings a smaller improvement of $0.9\%$ (from 65.0\% to 65.9\%) and $0.7\%$ (from 55.6\% to 56.3\%) on old and new classes, respectively.
This is expected since it concerns a more limited number of classes and exploits the model for the new classes that is being learned in the current step, that is still not completely reliable. 
Notice that  differently from background self-inpainting, knowledge self-inpainting do not affect labels of old classes. Since the learning progress tends to be more and more coarse, performing inpainting of replay images without  constraints  reduces  performances (see Table \ref{tab:paint_vs_mix}). In the Table, we also studied the effect on Equation \ref{eq:knowledge_inpainting} of adding a further constraint on the classes already present (in  column three). 



\begin{table}[]
\centering
\caption{Ablation Study: mIoU on PASCAL VOC2012 10-1  (disjoint).  \\ AL: Adversarial Learning\um{.} CTH: Per-class Thresholding}
\setlength{\tabcolsep}{3pt}
\label{tab:abl_all}
\begin{tabular}{c|c|c|c|c||cc|c} \hline
Replay & AL & CTH & \begin{tabular}[c]{@{}c@{}}Background\\ Inpainting\end{tabular} & \begin{tabular}[c]{@{}c@{}}Knowledge\\ Inpainting\end{tabular} & old  & new  & all  \\ \hline
\checkmark      &    &    &     &       & 46.5 & 33.6 & 51.6 \\ 
\checkmark      & \checkmark   &     &     &       & 60.1 & 45.4 & 54.1 \\ 
\checkmark      & \checkmark      & \checkmark   &   &    & 61.9 & 46.2 & 55.3 \\ 
\checkmark      &    & \checkmark   & \checkmark    & \checkmark   & 65.6 & 53.8 & 61.0 \\ 
\checkmark      & \checkmark    &    & \checkmark    & \checkmark   & 64.6 & 54.5 & 60.9 \\ 
\checkmark      & \checkmark    &  \checkmark  &     & \checkmark   & 64.0 & 45.4 & 56.1 \\ 
\checkmark      & \checkmark      & \checkmark   & \checkmark    &    & 65.0 & 55.6 & 61.6 \\ 
\checkmark & \checkmark & \checkmark & \checkmark & \checkmark  & \textbf{65.9} & \textbf{56.3} & \textbf{62.3} \\ \hline
\end{tabular}
\end{table}

\noindent \textbf{Image Selection Methods} 
Table \ref{tab:abl_is} shows the ablation studies concerning the proposed image selection strategies, \ie, the adversarial learning-based (AL) selection strategy and the CDF-based Threshold (CTH) selection strategy. Besides, the fixed threshold strategy (FTH) is also listed as a reference. FTH uses a fixed label size threshold: $25\%$ of the image area, fro the selection strategy.
The baseline strategy without any selection strategy leads to an accuracy of $51.6$. Notice that the self-inpainting techniques are not adopted here. 



\begin{table}[t]
\caption{Impact of keeping the currently existing label area unchanged.  
Setting: PASCAL VOC2012 10-1  (disjoint).}
\centering
\begin{tabular}{c|cc}
\hline
w/o \begin{tabular}[c]{@{}c@{}}Knowledge\\ Inpainting\end{tabular} & w/o Constraints  & w Constraints \\ \hline
61.6      & 59.3 & \textbf{62.3}   \\ \hline
\end{tabular}
\label{tab:paint_vs_mix}
\end{table}

\begin{table}[t]
\centering
\caption{Ablation study of the proposed image selection methods. Setting: PASCAL VOC2012 10-1  (disjoint).}
\label{tab:abl_is}
\begin{tabular}{ccc} \hline
w/o Selection      & \textbf{F}ixed \textbf{TH}reshold & \textbf{C}DF \textbf{TH}reshold \\ 
51.6               & 53.8            & 51.3          \\ \hline
\textbf{A}dversial \textbf{L}earning & AL+FTH           & AL+CTH         \\
54.1               & 53.4            & \textbf{55.3}         \\ \hline
\end{tabular}
\end{table}

Let's consider the AL first. By incorporating the AL filtering strategy, it brings an improvement of $2.5\%$ w.r.t the baseline, which indicates that the dataset with more accurately selected images is beneficial to the model. When separately using CTH, it causes a slight drop of $0.3\%$ in performance, while using FTH separately gains an improvement of $2.2\%$, which indicates that the replay images  have to contain a larger label area to preserve the knowledge.
When combining AL with FTH and CTH respectively, AL+CTH gains an improvement of $1.9\%$ w.r.t AL+FTH, proving that semantic content thresholding without considering the sizes of different objects may cause degradation of performance. 

\begin{figure*}[htbp]
\centering
\vspace{-0.15cm}
\includegraphics[trim=3cm 3cm 5cm 3cm, clip, width=.9\linewidth]{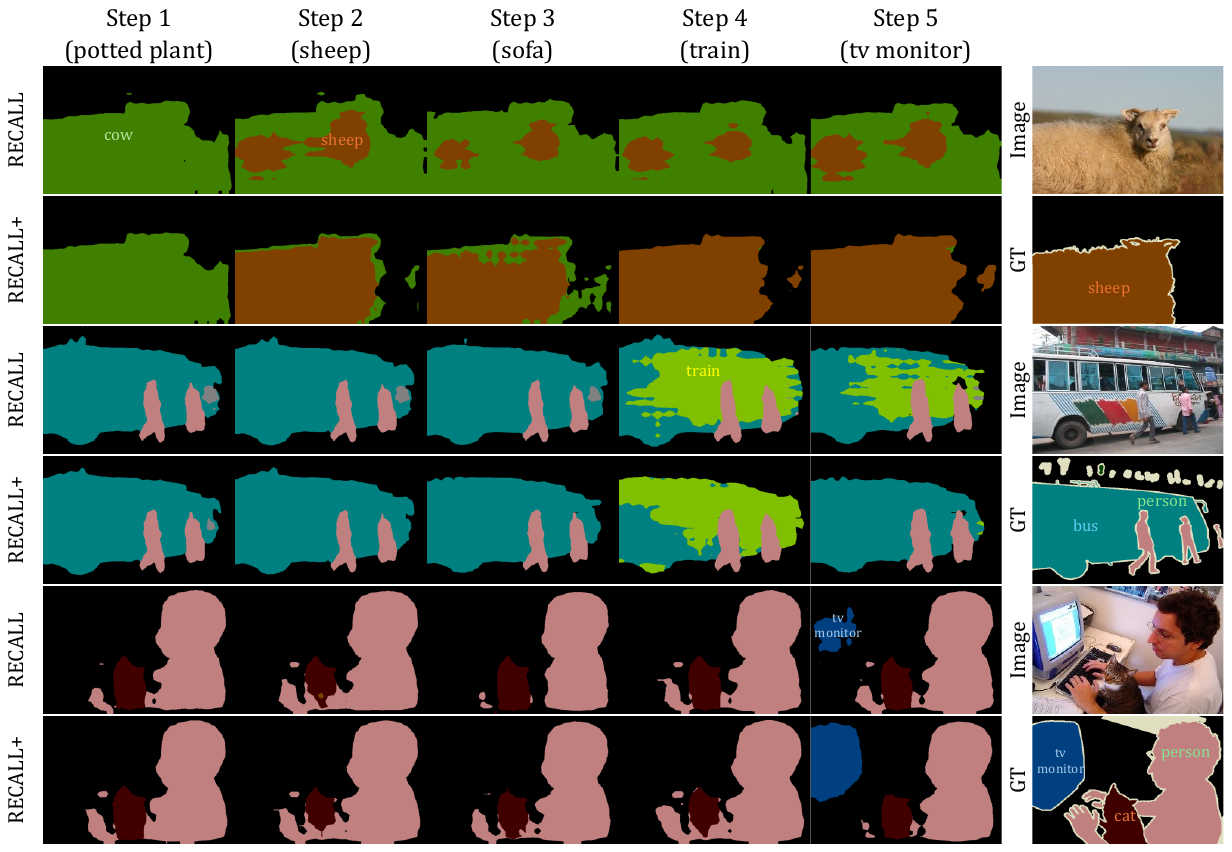}
\caption{Per-step prediction maps on disjoint incremental setup 15-1 on the Pascal VOC 2012 validation dataset.} 
\label{fig:qualitative_recall}
\vspace{-0.1cm}
\end{figure*}

\noindent \textbf{Per step prediction}
Fig. \ref{fig:qualitative_recall} visualizes the predictions for each step in the task-disjoint scenario 15-1. Initially, both models output equivalent predictions and tend to assign the unseen object with most similar label (see \textit{sheep}). However, RECALL quickly forgets the previous classes and becomes biased towards new classes. Differently, RECALL+ predictions are much more stable on old classes while learning new ones. 

\noindent \textbf{Web images numbers}
Here we present an ablation study on the adoption of a varying amount of web images. As shown in Table \ref{tab:abl_diff-num}, it can be observed that querying more web images leads to improved performance, although it also results in increased downloading time. We chose to adopt 10,000 web images because it achieves state-of-the-art (SOTA) performance compared to the other quantities. However, one could also opt for 5,000 web images to reduce time consumption.

\begin{table}[]
\caption{Ablation study of using different number of web images under disjoint 10-1.}
\centering
\begin{tabular}{c|cccc}
\hline
 Web image number & 1000 & 3000 & 5000 & 10000 \\ \hline
 Downloading time & 13 min & 42 min & 66 min & 133 min \\ \hline
mIoU & 59.5   & 60.3   & 61.5 & \textbf{62.3}   \\ \hline
\end{tabular}
\label{tab:abl_diff-num}
\end{table}

\section{Conclusions}
\label{sec:conclusions}
In this paper, we tackle  catastrophic forgetting by proposing a\fb[]{n efficient }
yet effective replay strategy using web-crawled images. 
We extracted from web images the most suitable for training using a selection strategy exploiting two main insights: an adversarial learning strategy, that aims at selecting images with statistics resembling the original training ones; and an analysis of the semantic prediction, which filters images whose semantic labels do not match the expected semantic content.
Besides, we extend the background self-inpainting technique to the web-downloaded images, further improving the performance.
The exhaustive experimental evaluation proves the effectiveness and robustness of the proposed approach. 

\section{Discussion and Future Research Directions} 
\label{sec:discussion}
Although the proposed approach gains satisfactory results on many incremental scenes, there are still some limitations. The first is that the discriminator has to be fine-tuned before each incremental step. One possible approach is to build a new domain discriminator that is trained only at the initial step and can filter out unseen classes without requiring fine-tuning. Another limitation is that the capability of preserving previous knowledge heavily depends on the helper decoders. The main model may struggle to preserve old knowledge effectively if the decoders cannot generate satisfactory pseudo-labels (as seen in the very challenging benchmark ADE20k). Additionally, new research will focus on exploring how to exploit unsatisfactory web-crawled data instead of just discarding it. 
Finally, the weak-supervision can also be considered as a strategy to replace pixel-wise labelling and tackle more challenging settings where only image-level labeling is available in the incremental steps. 

\bibliographystyle{IEEEtran}
\bibliography{strings,refs}






\vfill

\end{document}

%% file: sections/results_tables_1_2.tex
\begin{table*}[htbp]
\centering
\footnotesize
\setlength{\tabcolsep}{3pt}
\renewcommand{\arraystretch}{0.41}

\caption{Results on the Pascal VOC2012 dataset for different continual learning scenarios. 
$\dagger$: excerpted from \cite{maracani2021recall}, $\ddagger$: excerpted from the original paper in the available scenarios. $*$: background is included in the reported result. \\VOC: extra data from the training set WEB: extra data from the web.}

\begin{tabular}{l | l | c !{\vline width 2pt} cc|c|c | cc|c|c  !{\vline width 2pt} cc|c|c | cc|c|c } 
\toprule
\multicolumn{1}{c}{} & \multicolumn{1}{c}{} & \multicolumn{1}{c}{} & \multicolumn{8}{c}{\textbf{19-1}} & \multicolumn{8}{c}{\textbf{15-5}} \\

& & & \multicolumn{4}{c|}{\textbf{Disjoint}} & \multicolumn{4}{c !{\vline width 2pt}}{\textbf{Overlapped}}

&  \multicolumn{4}{c|}{\textbf{Disjoint}} & \multicolumn{4}{c}{\textbf{Overlapped}} \\[0.5ex]

\textbf{Method} & \textbf{Joint} & \begin{tabular}[c]{@{}c@{}}\textbf{Extra}\\ \textbf{Data}\end{tabular} & 1-19 & \phantom{6}20\phantom{6} & all &  $\Delta \downarrow$     & 1-19 & \phantom{6}20\phantom{6} & all & $\Delta \downarrow$
                & 1-15 & 16-20 & all & $\Delta \downarrow$     & 1-15 & 16-20 & all & $\Delta \downarrow$   \\
\midrule

$\text{FT}^\dagger$  & 75.4 & - & 35.2 & 13.2 & 34.2 & 41.2 & 34.7 & 14.9 & 33.8 & 41.6
     & 8.4 & 33.5 & 14.4 & 61.0 & 12.5 & 36.9 & 18.3 & 57.1 \\
     
$\text{S\&R}^\dagger$ & 75.4 & VOC & 55.3 & 43.2 & 56.2 & 19.2 & 54.0 & 48.0 & 55.1 & 20.3
      & 38.5 & 43.1 & 41.6 & 33.8 &  36.3 & 44.2 & 40.3 & 35.1 \\ 
\midrule

$\text{LwF}^\dagger$ \cite{li2017learning} & 75.4 & - & 65.8 & 28.3 & 64.0 & 11.4 & 62.6 & 23.4 & 60.8 & 14.6
                                    & 39.7 & 33.3 & 38.2 & 37.2 & 67.0 & 41.8 & 61.0 & 14.4 \\

$\text{LwF-MC}^\dagger$ \cite{rebuffi2017icarl} & 75.4 & VOC & 38.5 & 1.0 & 36.7 & 38.7 & 37.1 & 2.3 & 35.4 & 40.0
                                         & 41.5 & 25.4 & 37.6  & 37.8 & 59.8 & 22.6 & 51.0 & 24.4\\

$\text{ILT}^\dagger$ \cite{michieli2019incremental}  & 75.4 & - & 66.9 & 23.4 & 64.8 & 10.6 & 50.2 & 29.2 & 49.2 & 26.2
                                             & 31.5 & 25.1 & 30.0 & 45.4 & 69.0 & 46.4 & 63.6 & 11.8\\
                                             
$\text{CIL}^\dagger$ \cite{klingner2020class} & 75.4 & - & 62.6 & 18.1 & 60.5 & 14.9 & 35.1 & 13.8 & 34.0 & 41.4
                                       & 42.6 & 35.0 & 40.8 & 34.6 & 14.9 & 37.3 & 20.2 & 55.2\\

$\text{MiB}^\dagger$ \cite{cermelli2020modeling} & 75.4 & - & 69.6 & 25.6 & 67.4 & 8.0 & {70.2} & 22.1 & 67.8 & 7.6
                                          & 71.8 & 43.3 & 64.7 & 10.7 & {75.5} & 49.4 & 69.0 & 6.4\\

$\text{SDR}^\dagger$ \cite{michieli2021continual} & 75.4 & - & {69.9} & 37.3 & {68.4} & 7.0 & 69.1 & 32.6 & 67.4 & 8.0
                                           & {73.5} & 47.3 & {67.2} & 8.2 & {75.4} & 52.6 & {69.9} & 5.5  \\

PLOP${}^\ddagger$ \cite{douillard2021plop} & 77.4 & - &  $\text{75.4}^*$ & 37.4 & {73.5} & 3.9 & - & - & - & -
                                           & $\underline{75.7}^*$ & 51.7 & \underline{70.1} & \underline{7.3} & - & - & - & -  \\ 

REMI${}^\ddagger$ \cite{phan2022class} & 77.5 & - &  $\textbf{76.5}^*$ & 32.3 & \underline{74.4} & \underline{3.1} & - & - & - & -
                                           & $\textbf{76.1}^*$ & 50.7 & \underline{70.1} & 7.4 & - & - & - & -  \\ 

RCIL${}^\ddagger$ \cite{zhang2022representation} & 78.2 & - &  - & - & - & - & - & - & - & -
                                           & $\text{75.0}^*$ & 42.8 & 67.3 & 10.9 & $\textbf{78.8}^*$ & 52.0 & \textbf{72.4} & 5.8\\ 
SPPA${}^\ddagger$ \cite{lin2022continual} & 77.5 & - &  $\text{75.5}^*$ & 38.0 & 73.7 & 3.8 & $\text{76.5}^*$ & 36.2 & \underline{74.6} & \underline{2.9}
                            & $\text{75.3}^*$ & 48.7 & 69.0 & 8.5 & $\underline{78.1}^*$ & 52.9  & \underline{72.1} & \underline{5.4} \\ 
RBC${}^\ddagger$ \cite{zhao2022rbc} & 77.4 & - & $\underline{76.4}^*$ & 45.8 & \textbf{75.0} & \textbf{2.4} & $\underline{77.3}^*$ & \textbf{55.6} & \textbf{76.2} & \textbf{1.2}
                        & $\text{75.1}^*$ & 49.7 & 69.9 & 7.5 & $\text{76.6}^*$ & 52.8 & 70.9 & 6.5 \\ 
SSUL+AWT${}^\ddagger$ \cite{goswami2023attribution} & 77.4 & VOC &  - & - & - & - & - & - & - & -
                                           & - & - & - & - & $\text{78.0}^*$ & 50.2 & 71.4 & 6.0 \\ 

MiB+EWF${}^\ddagger$ \cite{xiao2023endpoints} & 78.2 & - &  - & - & - & - & $\textbf{77.9}^*$ & 6.7 & {74.5} & -
                                           & - & - & - & - & - & - & - & - \\ \midrule

RECALL \cite{maracani2021recall} & 75.4 & WEB & 65.0 & \underline{47.1} & 65.4  & 10.0 & 68.1 & \underline{55.3} & {68.6} & 6.8
           & 69.2 & \underline{52.9} & 66.3 & 9.1 & 67.7 & \underline{54.3} & 65.6 & 9.8 \\ 

\midrule

RECALL+ & 75.5 & WEB & {70.7} & \textbf{55.6} & {70.9} & 4.6 & {70.7} & \textbf{55.6} & {70.9} & 4.6
           & {73.6} & \textbf{60.4} & \textbf{71.2} & \textbf{4.3} & 73.6 & \textbf{60.4} & {71.2} & \textbf{4.3} \\ \midrule



\multicolumn{11}{c}{}\\[-1ex]
\multicolumn{1}{c}{} & \multicolumn{1}{c}{} & \multicolumn{1}{c}{} & \multicolumn{8}{c}{\textbf{15-1}} & \multicolumn{8}{c}{\textbf{10-10}} \\
& & & \multicolumn{4}{c}{\textbf{Disjoint}} & \multicolumn{4}{c !{\vline width 2pt}}{\textbf{Overlapped}}
& \multicolumn{4}{c}{\textbf{Disjoint}} & \multicolumn{4}{c}{\textbf{Overlapped}} \\[0.5ex]

\textbf{Method} & \textbf{Joint} & \begin{tabular}[c]{@{}c@{}}\textbf{Extra}\\ \textbf{Data}\end{tabular} & 1-10 & 11-20 & all & $\Delta \downarrow$    & 1-10 & 11-20 & all & $\Delta \downarrow$

                & 1-10 & 11-20 & all & $\Delta \downarrow$    & 1-10 & 11-20 & all & $\Delta \downarrow$   \\

\midrule
$\text{FT}^\dagger$ & 75.4 &- & 5.8 & 4.9 & 5.6 & 69.8  & 4.9 & 3.2 & 4.5 & 70.9
   & 7.7 & 60.8 & 33.0  & 42.4 & 7.8 & 58.9 & 32.1 & 43.3 \\

$\text{S\&R}^\dagger$  & 75.4 & VOC & 41.0 & 31.8 & 40.7 & 34.7 & 38.6 & 31.2 & 38.9 & 36.5
       & 25.1 & 53.9 & 41.7 & 33.7 & 18.4 & 53.3 & 38.2 & 37.2\\ 
\midrule

$\text{LwF}^\dagger$ \cite{li2017learning} & 75.4 &- & 26.2 & 15.1 & 23.6 & 51.8 & 24.0 & 15.0 & 21.9 & 53.5
                           & 63.1 & 61.1 & 62.2 & 13.2 & \textbf{70.7} & 63.4 & \underline{67.2} & \underline{8.2}\\

$\text{LwF-MC}^\dagger$ \cite{rebuffi2017icarl}& 75.4 & VOC & 6.9 & 2.1 & 5.7 & 69.7 & 6.9 & 2.3 & 5.8 & 69.6
                               & 52.4 & 42.5 & 47.7  & 27.7 & 53.9 & 43.0 & 48.7 & 26.7\\

$\text{ILT}^\dagger$ \cite{michieli2019incremental} & 75.4 &- & 6.7 & 1.2 & 5.4  & 70.0 & 5.7 & 1.0 & 4.6 & 70.8
                                   & \underline{67.7} & \underline{61.3} & \underline{64.7} & \underline{10.7} & 70.3 & 61.9 & 66.3 & 9.1 \\

$\text{CIL}^\dagger$ \cite{klingner2020class} & 75.4 &- & 33,3 & 15.9 & 29.1  & 46.3 & 6.3 & 4.5 & 5.9  & 69.5
                             & 37.4 & 60.6 & 48.4 & 27.0 & 38.4 & 60.0 & 48.7 & 26.7\\

$\text{MiB}^\dagger$ \cite{cermelli2020modeling} & 75.4 &- & 46.2 & 12.9 & 37.9 & 37.5 & 35.1 & 13.5 & 29.7 & 45.7
                                          & 66.9 & 57.5 & 62.4  & 13.0 & 70.4 & \underline{63.7} & \underline{67.2} & \underline{8.2}\\

$\text{SDR}^\dagger$ \cite{michieli2021continual} & 75.4 &- & 59.2 & 12.9 & 48.1 & 27.3 & 44.7 & 21.8 & 39.2 & 36.2
                                           & 67.5 & 57.9 & 62.9  & 12.5 & \underline{70.5} & \textbf{63.9} & \textbf{67.4} & \textbf{8.0}\\ 

PLOP${}^\ddagger$ \cite{douillard2021plop} & 77.4 &- & $\text{65.1}^*$ & 21.1 & 54.6 & 22.8 & - & - & - & -
                                           &  - & - & -& - & - & - & - & - \\ 

REMI${}^\ddagger$ \cite{phan2022class} & 77.5 &- & $\underline{68.3}^*$ & 27.2 & 58.5 & {19.0} & - & - & - & -
                                           & - & - & - & - & - & - & - & -\\ 

RCIL${}^\ddagger$ \cite{zhang2022representation} & 78.2 &- & $\text{66.1}^*$ & 18.2 & 54.7 & 23.5 & $\text{70.6}^*$ & 23.7 & 59.4 & 18.8
                                           & - & - & - & - & - & - & - & - \\ 

SPPA${}^\ddagger$ \cite{lin2022continual} & 77.5 &- & $\text{59.6}^*$ & 15.6 &  49.1 & 28.4 & $\text{66.2}^*$ & 23.3 & 56.0 & 21.5
                                           & - & - & - & - & - & - & - & - \\
RBC${}^\ddagger$ \cite{zhao2022rbc} & 77.4 &- & $\text{61.7}^*$ & 19.5 & 51.6 & 25.8 & $\text{69.5}^*$ & 38.4 & 62.1 & 15.3
                                       & - & - & - & - & - & - & - & - \\ 

SSUL+AWT${}^\ddagger$ \cite{goswami2023attribution} & 77.4 & VOC & - & - & - & - & $\underline{77.0}^*$ & 37.6 & \underline{67.6} & \underline{9.8}
                                           & - & - & - & - & - & - & - & - \\

MiB+EWF${}^\ddagger$ \cite{xiao2023endpoints} & 78.2 &- & - & - & - & - & $\textbf{77.7}^*$ & 32.7 & 67.0 & 11.2
                                           & - & - & - & - & - & - & - & -\\ \midrule

RECALL \cite{maracani2021recall} & 75.4 & WEB & 67.6 & \underline{49.2} & \underline{64.3} & \underline{11.1} & 67.8 & \textbf{50.9} & 64.8 & 10.6
           & 64.1 & 56.9 & 61.9 & 13.5 & 66.0 & 58.8 & 63.7 & 11.7 \\ 
\midrule

RECALL+ & 75.5 & WEB & \textbf{72.9} & \textbf{52.7} & \textbf{68.9} & \textbf{6.6} & {72.5} & \underline{50.6} & \textbf{68.1} & \textbf{7.4}
           & \textbf{70.6} & \textbf{63.6} & \textbf{68.2}  & \textbf{7.3} & 69.8 & {62.4} & \underline{67.2} & 8.3\\ \midrule

\multicolumn{11}{c}{}\\[-1ex]
\multicolumn{1}{c}{} & \multicolumn{1}{c}{} & \multicolumn{1}{c}{} & \multicolumn{8}{c}{\textbf{10-5}} & \multicolumn{8}{c}{\textbf{10-1}}\\
& & & \multicolumn{4}{c}{\textbf{Disjoint}} & \multicolumn{4}{c !{\vline width 2pt}}{\textbf{Overlapped}}
& \multicolumn{4}{c}{\textbf{Disjoint}} & \multicolumn{4}{c}{\textbf{Overlapped}} \\[0.5ex]
\textbf{Method} & \textbf{Joint} & \begin{tabular}[c]{@{}c@{}}\textbf{Extra}\\ \textbf{Data}\end{tabular} & 1-10 & 11-20 & all & $\Delta \downarrow$     & 1-10 & 11-20 & all & $\Delta \downarrow$
                
                & 1-10 & 11-20 & all & $\Delta \downarrow$     & 1-10 & 11-20 & all & $\Delta \downarrow$   \\

\midrule
$\text{FT}^\dagger$ & 75.4 &- & 7.2 & 41.9 & 23.7  & 51.7 & 7.4 & 37.5 & 21.7 & 53.7
   & 6.3 & 2.0 & 4.3 & 71.1   & 6.3 & 2.8 & 4.7 & 70.7\\

$\text{S\&R}^\dagger$  & 75.4 & VOC & 26.0 & 28.5 & 29.7 & 45.7 & 22.2 & 28.5 & 27.9 & 47.5
       & 30.2 & 19.3 & 27.3  & 48.1 & 28.3 & 20.8 & 27.1 & 48.3\\ 
\midrule

$\text{LwF}^\dagger$ \cite{li2017learning} & 75.4 &- & 52.7 & 47.9 & 50.4 & 25.0 & 55.5 & 47.6 & 51.7 & 23.7
                           & 6.7 & 6.5 & 6.6 & 68.8 & 16.6 & 14.9 & 15.8 & 59.6\\

$\text{LwF-MC}^\dagger$ \cite{rebuffi2017icarl} & 75.4 & VOC & 44.6 & 43.0 & 43.8 & 31.6 & 44.3 & 42.0 & 43.2 & 32.2
                               & 6.9 & 1.7 & 4.4 & 71.0 & 11.2 & 2.5 & 7.1 & 68.3\\

$\text{ILT}^\dagger$ \cite{michieli2019incremental} & 75.4 &- & 53.4 & 48.1 & 50.9 & 24.5 & 55.0 & 44.8 & 51.7 & 23.7
                                   & 14.1 & 0.6 & 7.5  & 67.9 & 16.5 & 1.0 & 9.1 & 66.3\\

$\text{CIL}^\dagger$ \cite{klingner2020class} & 75.4 &- & 27.5 & 41.4 & 34.1 & 41.3 & 28.8 & 41.7 & 34.9 & 40.5
                             & 7.1 & 2.4 & 4.9 & 70.5 & 6.3 & 0.8 & 3.6 & 71.8\\

$\text{MiB}^\dagger$ \cite{cermelli2020modeling} & 75.4 &- & 54.3 & 47.6 & 51.1  & 24.3 & 55.2 & 49.9 & 52.7 & 22.7
                                          & 14.9 & 9.5 & 12.3  & 63.1 & 15.1 & 14.8 & 15.0 & 60.4\\

$\text{SDR}^\dagger$ \cite{michieli2021continual} & 75.4 &- & 55.5 & 48.2 & 52.0  & 23.4 & 56.9 & {51.3} & 54.2 & 21.2
                                           & 25.5 & 15.7 & 20.8  & 54.6 & 26.3 & 19.7 & 23.2 & 52.2 \\ 

PLOP${}^\ddagger$ \cite{douillard2021plop} & 77.4 &- & - & - & - & - & - & - & - & -
                                           & $\text{44.0}^*$ & 15.5 & 30.5 & 46.9 & - & - & - & - \\ 

REMI${}^\ddagger$ \cite{phan2022class} & 77.5 &- & - & - & - & - & - & - & - & -
                                           & - & - & - & - & - & - & - & - \\ 

RCIL${}^\ddagger$ \cite{zhang2022representation} & 78.2 &- & - & - & - & - & - & - & - & -
                                           & $\text{30.6}^*$ & 4.7 & 18.2 & 60.0 & $\text{55.4}^*$ & 15.1 & 34.3 & 43.9 \\ 

SPPA${}^\ddagger$ \cite{lin2022continual} & 77.5 &- & - & - & - & - & - & - & - & -
                                           & - & - & - & - & - & - & - & - \\
RBC${}^\ddagger$ \cite{zhao2022rbc} & 77.4 &- & - & - & - & - & - & - & - & -
                                       & - & - & - & - & - & - & - & - \\ 

SSUL+AWT${}^\ddagger$ \cite{goswami2023attribution} & 77.4 & VOC & - & - & - & - & - & - & - & -
                                           & - & - & - & - & $\textbf{73.1}^*$ & 47.0 & \underline{60.7} & 16.7 \\

MiB+EWF${}^\ddagger$ \cite{xiao2023endpoints} & 78.2 &- & - & - & - & - & - & - & - & -
                                           & - & - & - & - & $\underline{71.5}^*$ & 30.3 & 51.9 & 26.3 \\ \midrule

RECALL \cite{maracani2021recall} & 75.4 & WEB & \underline{63.2} & \underline{55.1} & \underline{60.6} & \underline{14.8} & \underline{64.8} & \underline{57.0} & \underline{62.3} & \underline{13.1}
           & \underline{62.3} & \underline{50.0} & \underline{57.8} & \underline{17.6} & {65.0} & \underline{53.7} & \underline{60.7} & \underline{14.7} \\ 
\midrule

RECALL+ & 75.5 & WEB & \textbf{69.1} & \textbf{60.0} & \textbf{65.6} & \textbf{9.9} & \textbf{69.0} & \textbf{59.9} & \textbf{65.6} & \textbf{9.9}
           & \textbf{65.9} & \textbf{56.3} & \textbf{62.3} & \textbf{13.2} & {66.5} & \textbf{55.5} & \textbf{62.2} & \textbf{13.3}\\ 

\bottomrule

\end{tabular}

\label{tab:voc_1}
\end{table*}%


%% file: sections/table_ade.tex
\begin{table*}[]
\centering
\small
\setlength{\tabcolsep}{2pt} 
\caption{Results on the ADE20K dataset for different overlapped continual learning scenarios.
$\dagger$: excerpted from \cite{zhang2022representation}.$\ddagger$: excerpted from the original paper in the available scenarios.}
\label{tab:ade}
\begin{tabular}{l|l|c!{\vline width 2pt}cc|c|c!{\vline width 2pt}cc|c|c!{\vline width 2pt}cc|c|c!{\vline width 2pt}cc|c|c}
\toprule
\multicolumn{1}{c}{} & \multicolumn{1}{c}{} & \multicolumn{1}{c}{} & \multicolumn{4}{c}{{\textbf{100-50} (2 steps)}} &  \multicolumn{4}{c}{{\textbf{50-50} (3 steps)}}  &\multicolumn{4}{c}{{\textbf{100-10} (6 steps)}} &\multicolumn{4}{c}{{\textbf{100-5} (11 steps)}} \\
\textbf{Method}   & Joint & \begin{tabular}[c]{@{}c@{}}{Extra}\\ {Data}\end{tabular} & 1-100 & 101-150 & \textit{all} & $\Delta \downarrow$ &  1-50 & 51-150 &  \textit{all} & $\Delta \downarrow$ &  1-100 & 101-150 &  \textit{all} & $\Delta \downarrow$ & 1-100 & 101-150 &  \textit{all} & $\Delta \downarrow$ \\ \hline
$\text{ILT}^\dagger$\cite{michieli2019incremental} & 38.9 & - & 18.3 & 14.8 & 17.0 & 21.9 & 13.6 & 6.2 & 9.7 & 29.2 & 0.1 & 2.9 & 1.1 & 37.8 & 0.1 & 1.3 & 0.5 & 37.3\\
MiB$^\dagger$\cite{cermelli2020modeling}& 38.9 & - & 40.7 & 17.7 & 32.8 & 6.1 & 45.3 & 21.6 & 29.3 & 9.6 & 38.3 & 11.3 & 29.2 & 9.7 & 36.0 & 5.6 & 25.9 & 13 \\
PLOP$^\dagger$\cite{douillard2021plop}& 38.9 & - & {41.9} & 14.9 & 32.9 & 6.0 & {48.6} & 21.6 & 30.4 & 8.5 & {40.6} & 14.1 & 31.6 & 7.3 & {39.1} & 7.8 & 28.7 & 10.2 \\
{RCIL}$^\dagger$\cite{zhang2022representation}& 38.9 & - & {42.3}  & 18.8 & {34.5} & 4.4 & {48.3} & \underline{25.0} & {32.5} & 6.4 & 39.3 & 17.6 & 32.1 & 6.8 & 38.5 & 11.5 & {29.6} & 9.3 \\ 
REMI$^\ddagger$~\cite{phan2022class}& 39.0 & - & 41.6 & {19.2} & {34.1} & 4.9 & 47.1 & 20.4 & 29.4 & 9.6 & 39.0 & \underline{21.3} & {33.1} & 5.9 & 36.1 & \textbf{16.4} & 29.5 & 9.4\\
SPPA$^\ddagger$~\cite{lin2022continual}& 37.9 & - & \underline{42.9} & 19.9 & 35.2 & 2.7 & \textbf{49.8} & 23.9 & 32.5 & \underline{5.4} & \underline{41.0} & 12.5 & 31.5 & 6.4 & - & - & - &-d\\
RBC$^\ddagger$~\cite{zhao2022rbc}& 38.3 & - & \underline{42.9} & {21.5} & \underline{35.8} & \underline{2.5} & \underline{49.6} & \textbf{26.3} & \textbf{34.2} & \textbf{4.1} & 39.0 & \textbf{21.7} & \textbf{33.3} & \textbf{5.0} & - & - & -  & - \\

MiB+AWT$^\ddagger$\cite{goswami2023attribution}& 38.9 & - & 40.9 & \textbf{24.7} & 35.6 & 3.3 & 46.6 & - & \underline{33.5} & \underline{5.4} & 39.1 & - & \underline{33.2} & 5.7 & 38.6 & \underline{16.0} & 31.1 & 7.8 \\ 
MiB+EWF$^\ddagger$~\cite{xiao2023endpoints}& 38.9 & - & 41.2 & {21.3} & {34.6} & 4.3 & - & - & - & - & \textbf{41.5} & \underline{21.3} & \underline{33.2} & 5.7 & \textbf{41.4} & {13.4} & \textbf{32.1} & \underline{6.8} \\ \hline

RECALL+ & 38.2 & WEB & \textbf{43.4} & \underline{22.0} & \textbf{36.2} & \textbf{2.0} & 48.0 & {22.5} & {31.0} & 7.2 & {39.7} & {18.0} & {32.7} & \underline{5.5} & \underline{39.8} & {15.8} & \underline{31.6} & \textbf{6.6} \\ 
\bottomrule
\end{tabular}
\end{table*}